%% file: main.tex
\documentclass[10pt,twocolumn,letterpaper]{article}

\usepackage{wacv}              %

\input{preamble}

\definecolor{wacvblue}{rgb}{0.21,0.49,0.74}
\usepackage[pagebackref,breaklinks,colorlinks,allcolors=wacvblue]{hyperref}

\usepackage[capitalize]{cleveref}
\crefname{section}{Sec.}{Secs.}
\Crefname{section}{Section}{Sections}
\Crefname{table}{Table}{Tables}
\crefname{table}{Tab.}{Tabs.}

\def\wacvPaperID{421} %
\def\confName{WACV}
\def\confYear{2026}

\begin{document}
\title{Robust Context-Aware Object Recognition}
\input{authors}

\input{figures/teaser}

\begin{abstract}
In visual recognition, both the object of interest (referred to as foreground, \fg{}, for simplicity) and its surrounding context (background, \bg{}) play an important role. 
However, standard supervised learning often leads to unintended over-reliance on the \bg{}, known as shortcut learning of spurious correlations, limiting model robustness in real-world deployment settings.
In the literature, the problem is mainly addressed by suppressing the \bg{}, sacrificing context information for improved generalization.

We propose \methodabb{} --- Robust Context-Aware Object Recognition --- the first approach
that jointly achieves robustness and context-awareness without compromising either.
\methodabb{} treats localization as an integral part of recognition to decouple object-centric and context-aware modelling, followed by a robust, non-parametric fusion.
It improves the performance of both supervised models and \acp{vlm} on datasets with both in-domain and out-of-domain \bg{}, even without fine-tuning.
The results confirm that localization before recognition is now possible even in complex scenes as in ImageNet-1k.%
\vspace{-3ex}%
\footnote{The code will be made publicly available on GitHub.} 

\end{abstract}

\input{sections/intro}

\input{sections/related}
\input{sections/method}

\input{sections/implementation}

\input{sections/experiments}

\input{sections/conclusion}

\clearpage

{
    \small
    \bibliographystyle{ieeenat_fullname}
    \bibliography{main}
}

\clearpage

\appendix
\input{sections/app_embeddings}

\input{sections/app_details}

\input{sections/app_exps}

\end{document}

%% file: preamble.tex
\def\fg{{\scshape fg}}
\def\bg{{\scshape bg}}
\def\full{{\scshape full}}

\def\convnext{ConvNeXt}
\def\methodabb{RCOR}

\usepackage{graphicx}
\usepackage{amsmath}
\usepackage{amssymb}
\usepackage{booktabs}
\usepackage{algorithm}
\usepackage{algorithmic}

\usepackage[nolist,nohyperlinks]{acronym}

\usepackage{array} 

\usepackage{multirow}

\usepackage{adjustbox}

\usepackage{arydshln}

\usepackage[dvipsnames]{xcolor}

\usepackage{tikz}

\newcommand{\tikzxmark}{%
\tikz[scale=0.23] {
    \draw[line width=0.7,line cap=round] (0,0) to [bend left=6] (1,1);
    \draw[line width=0.7,line cap=round] (0.2,0.95) to [bend right=3] (0.8,0.05);
}}
\newcommand{\tikzcmark}{%
\tikz[scale=0.23] {
    \draw[line width=0.7,line cap=round] (0.25,0) to [bend left=10] (1,1);
    \draw[line width=0.8,line cap=round] (0,0.35) to [bend right=1] (0.23,0);
}}

\definecolor{darkgreen}{RGB}{0,128,0} %

\begin{acronym}
\acro{gt}[GT]{Ground Truth}
\acro{iou}[IoU]{Intersection over Union}
\acro{miou}[mIoU]{mean Intersection over Union}
\acro{ce}[CE]{Cross-Entropy}
\acro{vit}[ViT]{Vision Transformer}
\acro{sam}[SAM]{Segment Anything Model}
\acro{vlm}[VLM]{Vision-Language Model}
\end{acronym}

\DeclareMathOperator*{\argmax}{argmax}

%% file: authors.tex
\author{Klara Janouskova$^{1,2}$ \qquad Cristian Gavrus$^{1}$ \qquad Jiri Matas \\
Visual Recognition Group, Czech Technical University in Prague\\
{\tt\footnotesize \{klara.janouskova, gavrucri, matas\}@fel.cvut.cz} \\
{\footnotesize $^{1}$ equal contribution, $^{2}$ corresponding author} 
}

%% file: figures/teaser.tex
\twocolumn[
{
\renewcommand\twocolumn[1][]{#1}%
\maketitle
\begin{center}
\setlength{\tabcolsep}{0.1pt}
    
\begin{tabular}{>{\centering\arraybackslash}m{0.195\textwidth} >{\centering\arraybackslash}m{0.195\textwidth} >{\centering\arraybackslash}m{0.195\textwidth} >{\centering\arraybackslash}m{0.195\textwidth} >{\centering\arraybackslash}m{0.195\textwidth}}

\includegraphics[width=0.19\textwidth, height=3.2cm]{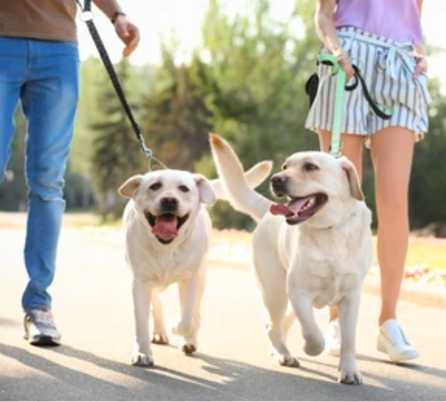}
&
\includegraphics[width=0.19\textwidth, height=3.2cm]{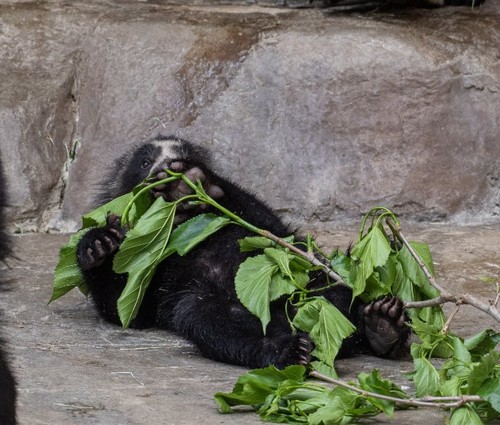}
&
    \includegraphics[width=0.19\textwidth, height=3.2cm]{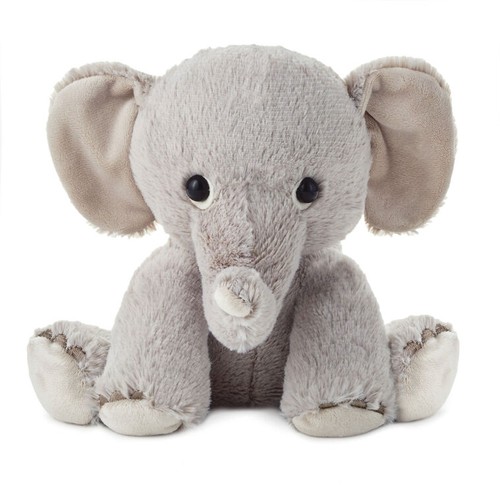}
    &
    \includegraphics[width=0.19\textwidth, height=3.2cm,]{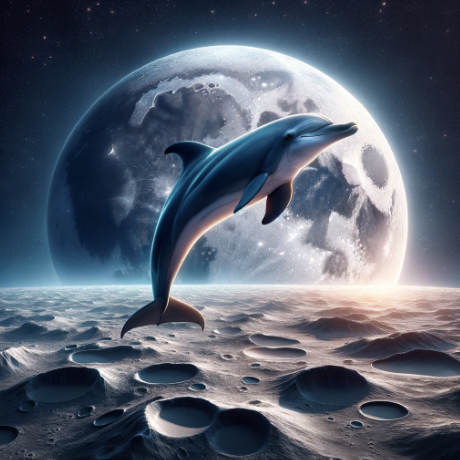} & 
    \includegraphics[width=0.19\textwidth, height=3.2cm]{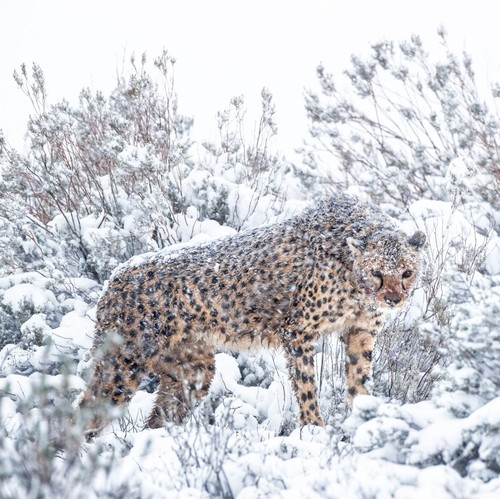} \\
    \small (a) \bg{}, the owners, critical for dog identification & 
    \small (b) the \bg{}  facilitates recognition &
    \small (c) \bg{} uninformative for classification &
    \small (d) generated \bg{} can be arbitrary &
    \small (e) long-tail \bg{}, not likely to appear during training
\end{tabular}
\begin{minipage}{\textwidth}
\vspace{0.5em}
\centering
\begin{tikzpicture}
    \draw[<->, thick] (0,0) -- (0.975\textwidth,0);
    \node[fill=white, inner sep=2pt, anchor=south] at (0.1\textwidth,-0.17) {\small context critical};
    \node[fill=white, inner sep=2pt, anchor=south] at (0.87\textwidth,-0.22) {\small context misleading};
\end{tikzpicture}
\end{minipage}
    \captionof{figure}{The complementarity of object (\fg{}) and context (\bg{}).
    The standard approach, \bg{} suppression, makes correct identification in (a) nearly impossible, and difficult in (b); the spectacled bear is the most herbivorous of all bear species, but its facial marks are partially occluded.  In generated content (d), any \fg{} can appear on any \bg{} as in ChatGPT 4o's response to ``a dolphin on the moon". Rare, even adversarial \bg{}s with possibly huge diversity hurt classification -- (e) shows a cheetah after a snowfall in South Africa, not a snow leopard.
   }
    \label{fig:teaser}
\end{center}
}
]

%% file: sections/intro.tex
\section{Introduction}
\label{sec:intro}
In standard object recognition, a neural network models the statistical distribution of object appearance in the training set based on the whole image.
This approach has been highly successful in i.i.d. settings, particularly with moderate to large-scale training data. 

As object recognition matured, analyses of its weaknesses \cite{singla2022salient,xiao2020noise,moayeri2022hard} revealed 
that supervised classifiers are particularly prone to unintended ``shortcut'' \cite{geirhos2020shortcut} over-reliance on the background (\bg{}) in the form of the so called spurious correlations \cite{ye2024spurious,izmailov2022feature},
where the model relies on particular \bg{} features instead of relevant object (\fg{}) properties.  
Moreover, \bg{} features  fail to generalize
to \bg{}s which are long-tail, i.e., rarely or never appearing in the training data, and to substantial \bg{} distribution shifts, not an uncommon situation. 
This seriously impacts model robustness in real-world deployment settings \cite{chen2023understanding,bhatt2024mitigating,lin2023spurious,beery2018recognition}.
This issue was later observed in \acp{vlm} as well, albeit to a lesser extent \cite{xu2025overcoming,wang2025sober}.

Recent methods address the problem by suppression of \bg{} features. 
They fall into two groups: the first emphasize \fg{} features during training \cite{bhatt2024mitigating,yang2024significant,chou2023fine,aniraj2023masking} by exploiting segmentation masks (often ground truth) or saliency maps,  the second alter the \bg{} distribution \cite{barbu2019objectnet,xiao2020noise,shetty2019not,ghosh2024aspire,wang2022clad} through image augmentation, including image generation, that introduces  less common \bg{}s.

\input{figures/vlms_context}

 The importance of both object appearance and the surrounding context in visual recognition has long been recognized across disciplines \cite{henderson1999high,chun1998contextual,torralba2003contextual,divvala2009empirical,oliva2007role}.
The nuanced role of \bg{}, as illustrated in \cref{fig:teaser}, is overlooked in recent recognition literature 
\cite{barbu2019objectnet,xiao2020noise,shetty2019not,ghosh2024aspire,wang2022clad}
where frequent co-occurrences of \fg{} and \bg{} are commonly dismissed as ``spurious correlations'' and considered harmful, a characterization we challenge as it ignores the important contribution of context to recognition.
Figure~\ref{fig:vlms_context} illustrates the problems of either over-relying on or dismissing contextual information, presenting two examples. In the top one, context enables correct recognition with CLIP \cite{radford2021learning}. In the bottom one, a misleading \bg{} causes an incorrect prediction despite a clear \fg{} object\footnote{CLIP-B predictions are from the online demo at \url{https://huggingface.co/spaces/merve/compare_clip_siglip}.}.

We propose the first object recognition approach, \methodabb{}, Robust Context-Aware Object Recognition, that jointly achieves robustness and context-awareness without compromising either. \methodabb{} uses class-agnostic localization as an integral part of the recognition process to decouple object-centric and context-aware representations, followed by a robust fusion.
We show it is possible to control the influence of the context, enabling models to use it when it is informative and to ignore it when it is misleading. Experiments confirm that zero-shot \fg{} localization as part of recognition is feasible with modern methods \cite{minderer2024scaling} even on challenging, multi-object scenes such as in the ImageNet \cite{deng2009imagenet,russakovsky2015imagenet} dataset and can be leveraged to improve object recognition on both in-domain (ID) and out-of-domain (OOD) data.

We experiment with both modern supervised models, \convnext-Tiny \cite{convnextv2} (zero-shot or fine-tuned on \fg{}), and a state-of-the-art \ac{vlm}, SigLIP2 \cite{tschannen2025siglip}, on a broad range of datasets. We consider datasets with ImageNet-1k classes, both 1. ID: ImageNet-1k \cite{russakovsky2015imagenet} validation set, its less noisy subset \cite{kisel2024flaws}, ImageNet-v2 \cite{recht2019imagenet} and the `common \bg{}' split of CounterAnimal \cite{wang2025sober} and 2. OOD: ImageNet-A \cite{hendrycks2021natural}, ImageNet-R \cite{hendrycks2021many}, ObjectNet \cite{barbu2019objectnet}, or the `rare \bg{}' split of the CounterAnimal dataset \cite{wang2025sober}. Additionally, we consider multiple fine-grained datasets: the widely adopted Stanford Dogs \cite{KhoslaYaoJayadevaprakashFeiFei_FGVC2011}, a synthetic, domain-generalization benchmark Spawrious \cite{lynch2023spawrious} and the domain-specific FungiTastic \cite{picek2024fungitastic}. For FungiTastic, we use BioCLIP \cite{stevens2024bioclip} instead of SigLIP2.

We first show that neither the standard context-aware \full{} nor the robustness-focused, object-centric \fg{} alone achieve both robustness to OOD \bg{}s and high IID performance; each trades performance in one of the cases for the other. We then show the proposed \methodabb{} achieves the best of both, \ie it has    performance close to $\max$(\fg{}, \full{}) across a wide range of scenarios,  often exceeding it. 

\methodabb{} offers additional advantages.  The decomposition opens new possibilities for \bg{} modelling, such as leveraging large pretrained models with strong representations, like DINO \cite{oquab2023dinov2} and CLIP \cite{radford2021learning}, or incorporating diverse data sources, such as tabular metadata representing the \bg{}. 

\noindent The contributions of this work are:
\begin{enumerate}[noitemsep,topsep=0pt]
    \item Introducing an object recognition approach, \methodabb{}, that disentangles object-centric (\fg{}) and context-aware (\full{}) representations, 
    being the first method enabling both robust and context-aware classification through a simple, interpretable, non-parametric fusion.
    \item Demonstrating that class-agnostic localization, namely with OWLv2 \cite{minderer2024scaling}, performs well enough to be integrated into object recognition pipelines, improving their robustness and performance.
    \item  Establishing zero-shot \fg{} as a strong baseline for \bg{} suppression, improving the performance of both supervised and \ac{vlm} classifiers on out-of-domain benchmarks (ImageNet-A/R, ObjectNet, CounterAnimal ‘rare \bg{}’), without fine-tuning.
    On the Spawrious \cite{lynch2023spawrious} domain generalization benchmark,
    it outperforms all state-of-the-art approaches which limit the \bg{} influence by modifying their training procedure, often relying on additional \bg{} annotations.
    \item The \methodabb{} fusion restores and even enhances ID performance while maintaining OOD performance. In contrast, the dominant robust recognition approach of \bg{} suppression represented by \fg{} hurts ID accuracy.
\end{enumerate}

%% file: figures/vlms_context.tex
\begin{figure}[bt]
    \centering
    \setlength{\tabcolsep}{0.05pt}
    \begin{tabular}
    {>{\raggedright\arraybackslash}m{0.4\linewidth} >{\raggedright\arraybackslash}m{0.6\linewidth}}

        \includegraphics[width=0.98\linewidth]{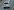} &
        \hfill
        \includegraphics[width=0.98\linewidth,trim={0cm 0.02cm 0cm 0.02cm},clip]{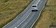} \\
        \small VLM: \hfill plug 46\% & \small \hfill  car 80\% \\
        \small \methodabb{} fusion: \small  \bf car & \\
         \\
         [-1em]
\includegraphics[width=0.98\linewidth]{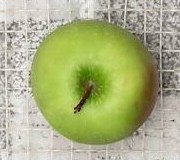} &
    \hfill    \includegraphics[width=0.98\linewidth,trim={0cm 0.4cm 0cm 0.4cm},clip]{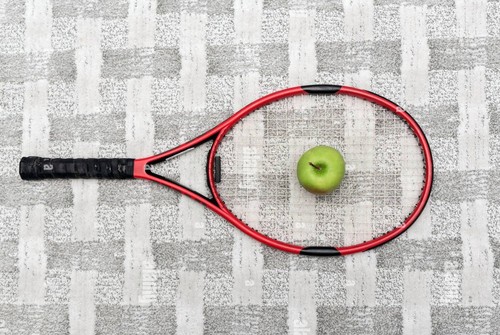} \\
        \small VLM: \hfill  apple 99 \% & \small \hfill  tennis ball 89 \%\\
        \small \methodabb{} fusion:  \small  \bf apple  & \\
    \end{tabular} 
    \caption{VLM (CLIP-B)
    -- zero-shot recognition with ground truth prompts and selected distractors. In the top example, recognition fails on the foreground (left, crop of a tight object bounding box). In the bottom, it fails on the full image (right).  The proposed robust fusion, \methodabb{}, is correct both times.
    }
    \label{fig:vlms_context}
\end{figure}

%% file: sections/related.tex
\section{Related work}
\label{sec:related}

\textbf{Complementary role of \fg{} and \bg{}.} 
Neuroscientific studies have shown that human perception integrates contextual cues to disambiguate objects and infer their identity in cluttered or ambiguous scenes \cite{henderson1999high,chun1998contextual}.
 Inspired by human vision, pioneering studies in object detection \cite{torralba2003contextual,divvala2009empirical,oliva2007role} emphasize the interdependence
between \fg{} and \bg{}. These works examine various types of contextual information such as co-occurrence statistics, spatial configurations, or scene-level constraints and demonstrate how contextual cues provide critical insights for recognition, sometimes more so than the object itself. 
Acharya et al.~\cite{acharya2022detecting} detect out-of-context objects through context provided by other objects within a scene, modelling co-occurrence through a Graph Neural Network (GNN).

In a recent study, Taesiri et al.~\cite{taesiri2024imagenet} dissect a subset of the ImageNet dataset \cite{russakovsky2015imagenet} into \fg{}, \bg{}, and \full{} image variants using ground truth bounding boxes.
A classifier is trained on each dataset variant, finding that the \bg{} classifier successfully identifies nearly 75\% of the images misclassified by the \fg{} classifier. Additionally, they demonstrate that employing zooming as a test-time augmentation markedly improves recognition accuracy.

Closely related to our approach, Zhu et al.~\cite{zhu2017object} advocate for independent modelling of \fg{} and \bg{} with post-training fusion. Unlike our method, which leverages recent advancements in zero-shot detection, their approach requires ground truth masks. A ground-truth-free approach is also proposed, but it consists of averaging 100 edge-detector-based bounding box proposals for each classifier \cite{zitnick2014edge}. This is not only extremely costly but also benefits heavily from ensembling, not necessarily showing benefits of independent modelling - most scenes in the evaluated ImageNet-1k dataset contain significantly less than 100 objects. The experiments are limited to a subset of a single, in-domain test dataset and weaker baselines (AlexNet \cite{krizhevsky2012imagenet}). In contrast, our work demonstrates the relevance and effectiveness of independent \fg{} modelling fused with context-aware prediction in modern settings, even in the context of large-scale vision-language models. Finally, unlike \cite{zhu2017object}, our work also focuses on robustness under \bg{} distribution shift.

Picek et al.~\cite{picek2024animal} investigate the role of \fg{} features and contextual metadata cues, such as time and location, in animal re-identification tasks. Unlike our general approach, their experiments specifically require the presence of ground-truth metadata, focus on niche applications and handcraft the \bg{} models.

Asgari et al.~\cite{asgari2022masktune} propose `MaskTune', a method which promotes the learning of a diverse set of features by masking out discriminative features identified by pre-training, without explicitly categorizing these features as \fg{} or \bg{}. The methods are quite different: (1) MaskTune is intended primarily for spurious correlations and selective classification datasets, while \methodabb{} applies to general datasets (2) MaskTune is designed only for supervised learning, while \methodabb{} applies to zero-shot settings as well 
(3) MaskTune involves training-time finetuning using xGradCAM to roughly mask images, while \methodabb{} is mostly an inference-time pipeline using a detector for precise localization 
(4) \methodabb{} was designed with interpretability in mind. 

\textbf{Background suppression.} Excessive reliance on \bg{} has a detrimental impact on classifier robustness to distribution shifts  \cite{moayeri2022comprehensive,xiao2020noise,bhatt2024mitigating,barbu2019objectnet,shetty2019not}.
In response, numerous strategies have been developed to mitigate this over-reliance by suppressing \bg{} during classification.
These methods typically involve regularizing classifier training to emphasize \fg{} features, either through the use of ground-truth segmentations or attention maps \cite{bhatt2024mitigating,yang2024significant,chou2023fine,aniraj2023masking}.
This enhances \fg{} representation but prevents the classifier from learning \bg{} cues that are necessary when \fg{} is ambiguous. Moreover, when \fg{}-\bg{} correlations are strong, reliance on attention maps for segmentation proves problematic, as the attention often highlights the \bg{} \cite{moayeri2022hard}.

Another group of methods involves training classifiers on images with manipulated or out-of-distribution backgrounds to reduce \bg{} dependency \cite{barbu2019objectnet,xiao2020noise,shetty2019not,ghosh2024aspire,wang2022clad}.
This technique results in complete disregard of \bg{} information or necessitates the modelling of \fg{}-\bg{} combinations for effective training, but it is not clear how to choose the optimal \bg{} distribution.

Deep-feature reweighting (DFR)\cite{kirichenko2023last} finetunes the last classifier layer to suppress \bg{} features. Unlike RCOR, which works even zero-shot and does not make assumptions about training data, DFR relies on a dataset without spurious correlations for training. 

Similarly to \methodabb{}, `CLIP with Guided Cropping' ~\cite{saranrittichai2024zeroshot} uses zero-shot open-vocabulary detection to focus on datasets with small objects. To note what sets \methodabb{} apart from Guided Cropping (GC) we mention: 
(1) GC focuses on predicting the \fg{} (\cite{saranrittichai2024zeroshot} emphasizing lack of context as a limitation) while \methodabb{} integrates the role of \fg{} and \bg{}. 
(2) For bounding box proposal used in evaluation: GC uses detection with text prompts associated to the top-$k$ classes - thereby errors in the initial classification can prevent correct object proposals from being considered at all. In contrast, \methodabb{} does not need text prompts for evaluation-detection and instead takes a class-agnostic approach based on objectness, decoupling object localization from classifier prediction. 
(3) GC focuses on zero-shot VLMs (like CLIP) classification, while we demonstrate the \methodabb{} method on both supervised models and VLMs. \cite{saranrittichai2024zeroshot}[sec A.4.2 and Table 5] acknowledges that GC does not attain optimal performance for supervised models in general.
(4) A favorable ImageNet comparison between RCOR and GC for CLIP-B is in \cref{tab:rw_comp:guided_crop}.

\textbf{\fg{}-\bg{} in other tasks.}
In the context of image segmentation, Mask2Former \cite{cheng2022masked} also adopts the \bg{} suuppression approach implemented by masking out \bg{} tokens in cross attention with queries inside the decoder to speed up convergence. The context is still incorporated in self-attention layers. A similar camouflage \bg{} approach is adopted in \cite{luo2023camouflaged}. More recently, Cutie \cite{cheng2024putting} extends this masked attention approach by separating the semantics of
the foreground object from the background for video object segmentation, focusing half of the object queries on the \fg{} and half on the \bg{}. While \fg{} only masked attention improves over standard attention, the \fg{}-\bg{} masked attention outperforms both, showing the importance of \bg{} information

Unlike in image classification, the field of image segmentation and tracking combines \bg{} suppression with contextual information, similarly to what we propose, but none adopts the independent \fg{} and context-aware \full{} modelling approach with robust fusion.

\textbf{Reliance on \bg{} in VLMs} is analyzed by \cite{wang2025sober} on a dataset of animals, where each animal is associated with two kinds of \bg{}, a `common' one (strong CLIP performance) and a `rare' one, where CLIP performance on the `rare' \bg{} drops significantly. Additionally, the lack of robustness to \bg{} shortcuts even in large-scale pretrained \acp{vlm} is confirmed by \cite{xu2025overcoming}.

\textbf{Zero-shot localization.} 
Recent advances in \acp{vlm} \cite{radford2021learning,liu2023grounding} and class-agnostic, promptable image detection and segmentation \cite{kirillov2023segment,ravi2024sam,ke2024segment,zhao2023fast} now facilitate zero-shot object localization of a wide range of objects without knowing their (fine-grained) class. This enables localization and effective \fg{}-\bg{} separation across a variety of image classification datasets.
Our methodology leverages these advances and seamlessly integrates robustness against unseen \bg{}s and utilization of the contextual information in \bg{}. 

%% file: sections/method.tex
\section{Method}
\label{sec:method}
\input{figures/full_pipeline}

The \methodabb{} method decouples the modelling of the object-centric \fg{} and the context-aware \full{} representation of an image and then combines them in a lightweight, interpretable module. It consists of three stages, see Figure \ref{fig:full_pipeline}: 1. Image decomposition to localize \fg{}, 2. independent \fg{} and \full{} appearance modelling, and 3. robust fusion. 

At inference time, the method relies solely on a class-agnostic object localizer $\mathcal{D}$, avoiding reliance on text prompts and category biases of traditional detectors. 
We assume $ \mathcal{D} $ outputs a bounding box for each object and its `objectness' confidence score.

\subsection{Object localization}
\label{sec:imdec}
The goal of this stage is to localize \fg{}, an image region $x_{\text{FG}}$ representing the target object. %
A key challenge in localizing $x_{\text{FG}}$ in general datasets (like ImageNet-1k) is the real-world scene complexity: images often contain multiple objects, and the one corresponding to the ground-truth label is not always the most prominent one. A class-agnostic detector will not identify the target object. For this reason, the localization stage may output multiple candidate  object regions and the process of object (bounding box) selection is detailed in Subsection \ref{sec:method:pred}.

 When training a classifier on \fg{} (optional but beneficial), we additionaly assume an open-vocabulary detector $ \mathcal{D'} $. In our experiments, both roles are covered by a single model, OWLv2, and $\mathcal{D} = \mathcal{D'}$.

\noindent\textbf{Inference-time localization.}
Given an image $x$,
the detector $ \mathcal{D}$ provides a set of candidate objects for $x_{\text{FG}}$: 
$  \{ (x_k, w_k) \}_{k \in \mathcal{K}} $
where each $x_k$
  is an image crop corresponding to a predicted bounding box and $w_k$ is an
  objectness score.

\noindent\textbf{Training-time localization (optional).}
While the method is agnostic to the choice of classifier (which can also be a standard supervised model trained on \full{} images without any fine-tuning or a \ac{vlm}), our experiments show training or fine-tuning the classifier on \fg{} images improves performance.  
Decoupling the \full{} and \fg{} training insures that the \fg{} classifier will not learn \bg{} shortcuts, which also improves interpretability.  %

Training-time boxes are obtained by a detector $\mathcal{D'}$ promptable with text (open-vocabulary) or images. 
For fine-grained datasets, we prompt with a text describing the dataset, \eg `dog' for dog species recognition, and select the box with the highest objectness score for training.
For general datasets, text prompts are replaced with per-class image queries from \cref{alg:class_embedding}. To generate image queries, bounding boxes are first generated with a per-class text-prompt based on the image ground truth label. The objectness scores of the top-2 boxes for each image then serve as input to \cref{alg:class_embedding}.  More details justifying this algorithm are in \cref{sec:emb_for_imagenet,sec:img_query_embeddings}.

\begin{algorithm}
\begin{algorithmic}[1]
\REQUIRE Set of images for a class, detector $ \mathcal{D}' $, param. $k$, the top-2 objectness scores $ s_1, s_2$ for each image.
\ENSURE Class embedding vector
\STATE Compute $ \gamma = \frac{s_1}{s_2} $ for all (or a subset of) class images (from the training set) and rank them by descending $ \gamma $.
\STATE (Optional)
Filter out images where top objectness box $ \neq $ top text box
\STATE Extract embeddings from each of the top-$k$ images top objectness region 
\STATE Take the mean of these top-$k$ embeddings to obtain the final class representation.
\end{algorithmic}
\caption{Image-conditioned Class Representation via Objectness Ratio}
\label{alg:class_embedding}
\end{algorithm}

\subsection{Prediction: candidate selection and fusion}
\label{sec:method:pred}
This stage generates \full{} and \fg{} predictions for the candidates from the previous stage. A single candidate region is then selected and fused with the \full{} image prediction via a simple, interpretable module.
In candidate selection and fusion, we prioritize robustness across evaluation datasets with differing distributions.

Suppose $ \Phi $ is a model (pre-trained or fine-tuned as in section \ref{sec:imdec}, or a \ac{vlm}) that outputs logit vectors $ \Phi(x) \in \mathbb{R}^C $, followed by a softmax activation $ \sigma $ to convert logits to per-class confidence. 
For each cropped image $ x_k $,we define the predicted class $ \hat{y}_k = \argmax_j \Phi(x_k)^{(j)}  $ and its confidence 
$ p_k = \sigma(\Phi(x_k))^{(\hat{y}_k)} $. 

\noindent\textbf{\fg{} crop selection and prediction.}
Among the multiple \fg{} crop candidates $x_k$'s predictions,
we choose the one maximizing the objectness-weighted confidence, that is
\begin{equation} 
\label{eq:weighted_confidence}
\hat{k} = \argmax_k (w_k p_k), \qquad \hat{y} = \hat{y}_{\hat{k}}
\end{equation}

\input{figures/weigh_boxes_fig}
 This strategy balances the generic, class-agnostic prominence of a region captured by the objectness score $w_k$ with the task-specific classification confidence $p_k$
  assigned by the model.
  The \textit{objectness weight} penalizes unclear, unfocused, or incomplete 
  objects as well as boxes that are over- or under-zoomed, see Fig.~\ref{fig:weight_conf_ablation} for examples. 
The \textit{classifier confidence} favours crops that are likely to match one of the target classes.

\noindent{\bf Fusion: \fg $\oplus$\full{}.}
We adopt an interpretable and non-parametric 
fusion approach to combine the robustness of  \fg{}  with the in-domain accuracy of context-aware \full{} models.
Let $p_F, \hat{y}_F$ be the confidence and prediction for 
\full{}.
The decision is then 
\begin{equation} 
\label{eq:fusion_def}
\hat{y} = 
\begin{cases}
\hat{y}_{\hat{k}} & \text{if } w_{\hat{k}} p_{\hat{k}} > w_1 p_F \\
\hat{y}_F & \text{otherwise}
\end{cases},
\end{equation}
assuming sorted weights (objectness scores) $ w_1 > w_2 > \cdots $. 
The intuition of assigning $w_1$ to the full image is that $w_1$ corresponds to the dominant object.

%% file: figures/full_pipeline.tex
\begin{figure}
        \includegraphics[trim={0 0cm 0 0cm},clip,width=\linewidth]
            {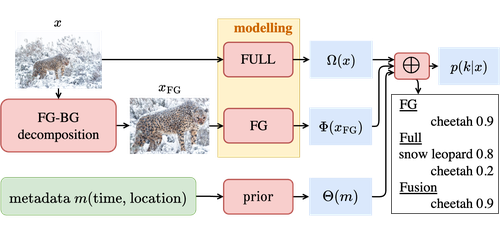}
        \caption{The proposed approach to robust context-aware recognition
     proceeds in three stages: (1) decomposition of image $x$ into \fg{} and \bg{} by zero-shot class-agnostic detection,
     (2) independent modelling of the \fg{} and the context-aware \full{} (original image), which also serves as a fallback option when detection fails, and (3) fusion that robustly combines the representations from stage (2) to form the output prediction $p(k|x)$.}
        \label{fig:full_pipeline}
\end{figure}

%% file: figures/weigh_boxes_fig.tex
\begin{figure}
\centering
\includegraphics[width=\linewidth]{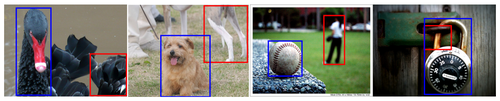}
    \caption{Localisation --- the role of objectness. Blue crops (maximising weighted confidence) lead to correct predictions, red crops (maximising unweighted confidence) lead to incorrect predictions, representing incomplete, unfocused or over-zoomed regions.}
    \label{fig:weight_conf_ablation}
\end{figure}

%% file: sections/implementation.tex
\section{Experimental Setup}
\label{sec:impl}
We provide two sets of experiments: (1) in the standard supervised training setup (2) using large-scale pretrained \acp{vlm} in a zero-shot recognition setup.
Additional details concerning the datasets and models are in the Appendix.

We evaluate the models on \full{} images in the standard manner and on \fg{} crops predicted by method \eqref{eq:weighted_confidence}, which then lead to \fg $\oplus$\full{} fusion predictions \eqref{eq:fusion_def}. In the case of the FungiTastic dataset, we assign the weight $ w=1$ to \full{} in \eqref{eq:fusion_def} instead of $ w_1 $, a choice determined on the validation set. 

\noindent\noindent\textbf{Evaluation metrics.} 
We report the most widely adopted \emph{total accuracy} metric
for most datasets. 
For the highly imbalanced FungiTastic, macro-averaged accuracy (mean of per-class accuracies) is reported.  
The ObjectNet dataset is evaluated with the multilabel \emph{ReAL accuracy}.

\subsection{Datasets with ImageNet-1k Classes}
\label{subsec:i1k_datasets}
We consider a wide range of general image classification evaluation datasets sharing the label space with ImageNet-1k (IN-1k)~\cite{russakovsky2015imagenet}. 
We group the datasets into two evaluation categories: 1. \emph{In-domain datasets}, where the contextual information (\bg{}) typically aids recognition, and domain shift is limited; and 2. \emph{Out-of-distribution datasets}, where the \bg{} is misleading, adversarial, or shifts significantly. This division, while defined relative to IN-1k, extends in part to vision-language models as well.

\textbf{In-Domain datasets (ID)} datasets are: 
ImageNet1k~\cite{russakovsky2015imagenet} (standard benchmark) and its `clean' subset with less noisy labels~\cite{kisel2024flaws}, 
ImageNetV2~\cite{recht2019imagenet} (generalization test, matched class distribution), 
Hard ImageNet (HIN)~\cite{moayeri2022hard} (IN1k subset with strong \fg{}-\bg{} correlation),
CounterAnimal-Common subset~\cite{wang2025sober} (Animals from iNaturalist on `common' \bg{}s).

 \textbf{Out-of-Domain datasets (OOD)} are:
 ImageNet-A~\cite{hendrycks2021natural} (natural adversarial samples, classification errors),
ObjectNet~\cite{barbu2019objectnet} (controlled \bg{}/viewpoint/rotation),
ImageNet-R~\cite{hendrycks2021many} (artistic/abstract renderings, distribution shift),
CounterAnimal-Rare subset~\cite{wang2025sober} (`rare' \bg{}s)
.

\subsection{Fine-grained Datasets}
\label{subsec:fine-grained_datasets}
To show broader applicability and highlight special cases, we consider:  
\textbf{FungiTastic (Fungi)}, \cite{picek2024fungitastic} (a challenging fungi species dataset with complex \fg{}-\bg{} relationships),
\textbf{Spawrious (Spaw)}\cite{lynch2023spawrious} (a synthetic dog-breed dataset 
where each class is associated with a specific \bg{} type  and the \bg{} distribution changes in the test set), a very similar but more widely adopted \textbf{Waterbirds} \cite{sagawa2019distributionally},
\textbf{Stanford Dogs (Dogs)}\cite{KhoslaYaoJayadevaprakashFeiFei_FGVC2011} (a dataset where the \bg{} plays no obvious role),

For {Dogs} and Spaw we reserve 15 \% of the training set for validation. For ImageNet-1k, we adopt the official validation set as the test set, a common practice in the literature.

\subsection{Generating bounding boxes}
\label{susbsec:bbox_gen}
We adopt the OWLv2 \cite{minderer2024scaling} detector \footnote{google/owlv2-large-patch14-ensemble}
\cite{wolf2020transformers}:
both as a class-agnostic detector (through its objectness head) and as an open-vocabulary 
detector from \cref{sec:method} and \cref{sec:app_loc}. 

\noindent\noindent\textbf{Bounding boxes for evaluation.}
For all test datasets, 
we collect bounding boxes and objectness scores for all images as described in Sec. \ref{sec:imdec}. We always include the highest scoring box and, additionally, all the boxes with objectness scores $>0.2$. 
The threshold value was not optimized  
since \cref{eq:weighted_confidence} automatically penalizes boxes with low score.

\noindent\noindent\textbf{Bounding boxes for training (optional).}
In the case of the fine-grained datasets, we prompt OWLv2 with a text describing the dataset: `dog' (Dogs, Spaw) or `mushroom, fungi' (FungiTastic).
For ImageNet-1k, text prompts are replaced with image queries obtained from Algorithm \ref{alg:class_embedding} (incl. step 2), with $ k=20 $, see App. \ref{sec:emb_for_imagenet} for more details.

\subsection{Supervised classification} \label{exp:setup}

\noindent\noindent\textbf{Models.}
We use a pretrained \convnext V2-Tiny model \footnote{convnextv2\_tiny.fcmae\_ft\_in1k} provided by the timm ~\cite{rw2019timm}. Its modern architecture is similar in size to ResNet-50, but achieves higher accuracy \cite{woo2023convnext}. 

For the fine-grained datasets we also finetune as follows. For FungiTastic begin with the same checkpoint\footnotemark[\value{footnote}]. 
For Spawrious we finetune a ResNet-50 model to compare with previous works. 
Since StanfordDogs is derived from ImageNet, but with much fewer samples per class,
we begin with a checkpoint  that was not pretrained on ImageNet \footnote{While most reported results on StanfordDogs \cite{dogs_leaderboard} use IN1k-pretrained models, this raises concerns about overlap with the test set and an uneven advantage, given ImageNet's significantly larger number of dog images.}.

 The training details are in Table \ref{tab:hyperparams2} in Supplementary.

\noindent\noindent\textbf{\fg{} training (optional).} For the \fg{} model we use either: (1) the same pretrained model used for full images (denoted \fg{}) or (2) we further fine-tune on the image crops generated from the training sets as described in Sec~\ref{sec:imdec} - then we denote the prediction by {\bf\fg{}$^+$} in the experiments and tables. 

In regards to \fg{} training for the fine-grained datasets we refer to Table \ref{tab:hyperparams2} again.

For ImageNet we largely follow the official ~\cite{convnextv2}
recipe  for training and augmentation, see ~\cref{tab:hyperparams} in Supplementary. 

The default crops 
~\cite{convnextv2, rw2019timm} (random-crop for training, center-crop for testing) are removed in our experiments.

\subsection{Vision-Language Models}
We adopt the state-of-the-art SigLIP2 \cite{tschannen2025siglip} (so400m-patch14-256) for all datasets with the exception of the FungiTastic, where evaluating general-purpose models is not meaningful
and we evaluate BioCLIP \cite{stevens2024bioclip} instead.
\fg{} inputs are padded to a square to preserve aspect ratio and avoid introducing unnecessary contextual information.

For each class $c$ with a text representation $t_c$, an embedding of `A photo of a $t_c$' is given by the text encoder, serving as the class prototype.
Each image is then classified based on the nearest class prototype to the image embedding.

\noindent\noindent\textbf{Text prompts.} 
The zero-shot performance of a \ac{vlm} is highly dependent on the per-class text prompts. The prompts vary between works, resulting in different baseline performance. We adopt the text prompts of \cite{kisel2024flaws}, which show state-of-the-art performance on ImageNet-1K \cite{russakovsky2015imagenet}.

%% file: sections/experiments.tex
\section{Results}
\label{sec:experiments}

\input{tables/in1k}
\input{tables/rw_comp/spaw_rw_comp}

\input{tables/summary}

\input{tables/rw_comp/guided_crop}
\input{tables/rw_comp/waterbirds}

The goal is to leverage contextual information when the object-centric prediction is uncertain (case 1, in-domain) while maintaining robustness to atypical or adversarial \bg{}s (case 2, out-of-domain).  We show that neither the standard context-aware \full{} nor the robustness-focused, object-centric \fg{} alone achieve both goals; each trades performance in one of the cases for the other. The proposed robust fusion matches the best of both, \ie it has performance close to max(\fg{}, \full{}) across a wide range of scenarios,  often exceeding it, see \cref{fig:app:exps:tradeoff}, \cref{ap:ablations}.

\subsection{Results on datasets with IN-1k classes}
Results for both supervised models and \acp{vlm} on ImageNet-derived evaluation datasets are reported in  \cref{tab:in1k}.

\noindent\textbf{Case 1: In-domain performance.}
The supervised \convnext evaluated on \fg{} suffers a performance drop w.r.t. \full{} on multiple of the datasets, lacking the contextual information of \bg{}. 

For SigLIP2, the performance drop on \fg{} inputs is consistent across all the evaluation datasets. The drop is more pronounced. A possible reason: for imprecise localization, the more general \ac{vlm} can not rely on the contextual cues in the wrong \fg{} crop as much as the supervised model does (as explained in Ablation on GT prompting).%

For both \convnext and SigLIP2, the proposed fusion not only recovers the performance lost by \fg{}, but also outperforms the standard \full{} approach. Part of this improvement can also be attributed to ensembling.
While the improvements of the fusion (\fg{} $\oplus$ \full) over \full{} are modest, this is expected, given that \full{} already incorporates the contextual information from the \bg{}.

The only dataset where \methodabb{} underperforms is Hard~ImageNet, see \cref{tab:in1k}. 
As an ablation with ground-truth guided localization later shows, see \cref{subsec:res:abl}, 
the localization pipeline is the limiting factor.
While the class-agnostic localization pipeline results in a 1.3 \% drop in accuracy,
when using GT-label prompts to localize \fg{}, 
the performance increases by 0.8 \% and 4.14 \%
for 
\fg{}$^+_{\text{GT}}$ 
$\oplus$ \full{}
and \fg{}$^+_{\text{GT}}$ $\oplus$ \full{}, 
respectively.

\noindent\textbf{Case 2: Out-of-domain performance.}
The supervised \convnext model benefits from \bg{} removal (evaluation on \fg{}) across all the datasets. The most notable improvements can be observed on ImageNet-A and ObjectNet with an increase by 19.6\% and 13.6\% compared to \full{}, respectively. These datasets were designed to contain unusual or even adversarial \bg{}s. We can further observe that our robust addition of context maintains the performance, or only very slightly decreases it, still outperforming standard evaluation on \full{} images by a large margin.

\noindent\textbf{Impact of fine-tuning on \fg{}.}
We investigate the impact of fine-tuning the standard ImageNet-pretrained \convnext{} on \fg{} cropped images in \cref{tab:in1k}. See Sec. \ref{exp:setup}, \ref{sec:imdec}. This model is denoted as $\text{\fg{}}^+$. 
An improvement is observed compared to the standard model evaluated on \fg{} on all OOD datasets. Notably, the performance on ImageNet-A improves from 29.99\% for the standard model to 37.95 \% for the fine-tuned one. The improvement is also reflected in the fusion results.
For the \fg{} image fine-tuned \convnext, the performance drop on in-domain data is stronger than for the \fg{} model. 

\noindent\textbf{Comparison to Guided Cropping \cite{saranrittichai2024zeroshot}.} \cref{tab:rw_comp:guided_crop} provides a comparison to \cite{saranrittichai2024zeroshot} using CLIP (ViT-B/32). \methodabb{} is shown to substantially outperform GC (and by a large margin on IN-A) on all datasets except IN-R, likely due to \methodabb{} being more context-aware and class-agnostic. Baselines differ most likely due to a difference in text-promtps.

\subsection{Ablations}
\label{subsec:res:abl}
\cref{ap:ablations} provides various ablations
, demonstrating the broad applicability of the 
fusion and localization method.

\noindent\textbf{(1) The role of the objectness weights.}
The \fg{} region selection and prediction (\cref{eq:weighted_confidence}, \eqref{eq:fusion_def}) 
rely on the objectness weights $ \{ w_k \}_k$.  
In  \cref{tab:noweights} App. \ref{ap:ablations}, 
 removal of the weights  is investigated, \ie, we evaluate the performance based on maximum classifier confidence only.
The objectness-weighted prediction performs better
across all datasets.

\noindent\textbf{(2) Different multi-box resolution strategies.} 
  In \cref{ap:ablations}, \cref{tab:siglip2_in1k_abl} we compare different resolution techniques to deal with multi-object images. The proposed method (\cref{eq:weighted_confidence}, \eqref{eq:fusion_def})  is compared to highest-score, union of all boxes and max-area multi-object resolution strategies. The results show that 
  \methodabb{} is stable across all scenarios.

\noindent\textbf{(3) Oracle (GT class) prompt localization.} 
In \cref{tab:gtprompts} we evaluate the \fg{} models on crops obtained using GT prompts \ie applying the (oracle) procedure "Training-time localization" (\cref{sec:method}) to the test sets. The results show potential for improvement over \full{} prediction under ideal localization.

\subsection{Fine-grained datasets experiments}
\noindent\textbf{Fine-grained datasets.}
Experiments with both supervised models and \acp{vlm} are reported in \cref{tab:finegrained}. 

\noindent\textbf{Stanford Dogs.}
For the supervised \convnext, we observe a significant improvement from \full{} to \fg{}$^+$ by 5.1\%, and an additional small improvement from \fg{}$^+$ to \fg{}$^+ \oplus $\full{} of 0.7\%. For the \ac{vlm}, no such effect is observed and the \fg{} performance drops a little, but the model still benefits from fusion, improving over \full{} by 0.8\%. 

\noindent\textbf{Spawrious.}
The \convnext{} experiment shows an adversarial situation for context-aware classifiers, where supervised models overfit to \bg{}, without learning strong, generalizable \fg{} features.
First notable thing is the underperformance of the \full{} model, reaching an accuracy of less than 40 \%. We observed that performance varies a lot among different checkpoints, as validation on a saturated dataset is not meaningful and different checkpoints exhibit different levels of \bg{} overfitting. The benefits of \fg{}$~^+$ are clear, improving over \full{} by over 56 \%. The fusion results still show some remaining trade-off of \methodabb between robustness and in-domain accuracy.

\noindent\textbf{FungiTastic.} We can observe a significant accuracy drop from \full{} to \convnext{} \fg{}$^+$ (-2.6 \%) and BioCLIP \fg{} (-6.8 \%). On one hand, the localization step is harder and cruder for this dataset. On the other hand, this could be explained by fungi species being strongly associated with certain environmental conditions reflected in the \bg{}, as well as information about co-occurrence, where multiple instances of the same or related specimens appear together. The effect is more significant for BioCLIP, which can be explained by the supervised \fg$^+$ model being trained on \fg{} while the BioCLIP \ac{vlm} may be relying on \bg{} shortcuts more. Fusion slightly improves performance for both supervised and \ac{vlm} models.

\subsection{Additional experiments}
\noindent\textbf{Comparison to domain generalization methods.}
To provide a fair comparison of the \methodabb{} to previous domain generalization methods, we provide results of Resnet50 classifiers on Spawrious \cite{lynch2023spawrious} \cref{tab:spaw:r50} and
\cref{tab:waterbirds}. The results establish \fg{}$+$ as superior (Spawrious) or competitive (Waterbirds) to other \bg{} suppression approaches, often relying on additional data such as group annotations. 
\fg{}$+ \oplus $ \full{} remains competitive, with only slightly deteriorated performance comapred to \fg{}, outperforming \full{} by a large margin. Note that \cite{noohdani2024decompose} and \cite{fan2021group} do not use the corrected Waterbirds \cite{asgari2022masktune} and the results are not directly comparable.

\textbf{State-of-the-art large supervised model} results are shown in \cref{ap:ablations}, \cref{tab:methods_vs_datasets_eva} 
\textbf{Alternative context model}
experiments are provided on the FungiTastic in \cref{ap:ablations}, \cref{tab:fungi_meta}. A  \textbf{comaprison to DFR} \cite{kirichenko2023last} with ResNet50 is in \cref{ap:ablations}, \cref{tab:dfr_shape_comp}, highlighting the strenghts and competitiveness of \methodabb{}.

%% file: tables/in1k.tex
\begin{table*}[ht]
\small
\setlength{\tabcolsep}{3.2pt}
\centering
\begin{tabular}{c l rrrrr rrrrr}
\toprule
 & method $\downarrow$  & \multicolumn{5}{c}{\textbf{in-domain:  \bg{} informative, no domain shift}} & \multicolumn{5}{c}{\textbf{out-of-domain: \bg{} uninformative or adversarial}} \\
\cmidrule(l){3-7}
\cmidrule(l){8-12}
    
& datasets $\rightarrow$ 
& 
\multicolumn{1}{c}{  IN-1K:Val} & 
\multicolumn{1}{c}{  IN-1K:Clean} & 
\multicolumn{1}{c}{  IN-V2} & 
\multicolumn{1}{c}{  Hard IN} & 
\multicolumn{1}{c}{  Animal-C} & 
\multicolumn{1}{c}{Animal-R} & 
\multicolumn{1}{c}{IN-A} & 
\multicolumn{1}{c}{  Object-Net} & 
\multicolumn{1}{c}{  IN-R} \\
\midrule

\multirow{6}{*}{\rotatebox[origin=c]{90}{{\scriptsize\textbf{\convnext}}}}
& \full{} & 82.35 & 93.12 & 70.97 & 81.33 & 87.26 & 71.62 & 10.36 & 25.70 & 33.89 \\
& \fg{} 
& {\scriptsize\textcolor{Red}{-0.12}} 82.23 
& {\scriptsize\textcolor{Red}{-0.24}} 92.88 
& {\scriptsize\textcolor{ForestGreen}{+0.96}} 71.93 
& {\scriptsize\textcolor{Red}{-4.93}} 76.40 
& {\scriptsize\textcolor{ForestGreen}{+1.89}} 89.15 
& {\scriptsize\textcolor{ForestGreen}{+6.02}} 77.64 
& {\scriptsize\textcolor{ForestGreen}{+19.63}} 29.99 
& {\scriptsize\textcolor{ForestGreen}{+13.55}} 39.25 
& {\scriptsize\textcolor{ForestGreen}{+2.64}} 36.53 \\
& \fg{}$\oplus$\full{} 
& {\scriptsize\textcolor{ForestGreen}{+1.03}} 83.38 
& {\scriptsize\textcolor{ForestGreen}{+0.74}} 93.86 
& {\scriptsize\textcolor{ForestGreen}{+2.01}} 72.98 
& {\scriptsize\textcolor{Black}{+0.00}} 81.33 
& {\scriptsize\textcolor{ForestGreen}{+2.70}} 89.96 
& {\scriptsize\textcolor{ForestGreen}{+6.07}} 77.69 
& {\scriptsize\textcolor{ForestGreen}{+15.28}} 25.64 
& {\scriptsize\textcolor{ForestGreen}{+12.16}} 37.86 
& {\scriptsize\textcolor{ForestGreen}{+2.94}} 36.83 \\
& \fg{}$^+$ 
& {\scriptsize\textcolor{Red}{-0.87}} 81.48 
& {\scriptsize\textcolor{Red}{-0.97}} 92.15 
& {\scriptsize\textcolor{ForestGreen}{+0.58}} 71.55 
& {\scriptsize\textcolor{Red}{-9.06}} 72.27 
& {\scriptsize\textcolor{ForestGreen}{+2.56}} 89.82 
& {\scriptsize\textcolor{ForestGreen}{+7.12}} 78.74 
& {\scriptsize\textcolor{ForestGreen}{+27.59}} 37.95 
& {\scriptsize\textcolor{ForestGreen}{+13.62}} 39.32 
& {\scriptsize\textcolor{ForestGreen}{+3.95}} 37.84 \\
& \fg{}$^+\oplus$\full{} 
& {\scriptsize\textcolor{ForestGreen}{+0.97}} 83.32 
& {\scriptsize\textcolor{ForestGreen}{+0.66}} 93.78 
& {\scriptsize\textcolor{ForestGreen}{+1.89}} 72.86 
& {\scriptsize\textcolor{Red}{-2.13}} 79.20 
& {\scriptsize\textcolor{ForestGreen}{+2.78}} 90.04 
& {\scriptsize\textcolor{ForestGreen}{+6.32}} 77.94 
& {\scriptsize\textcolor{ForestGreen}{+21.45}} 31.81 
& {\scriptsize\textcolor{ForestGreen}{+12.68}} 38.38 
& {\scriptsize\textcolor{ForestGreen}{+4.28}} 38.17 \\
\cmidrule{2-2}
& {\sc CenterCrop} 
& {\scriptsize\textcolor{ForestGreen}{+0.54}} 82.89 
& {\scriptsize\textcolor{ForestGreen}{+0.23}} 93.35 
& {\scriptsize\textcolor{ForestGreen}{+1.32}} 72.29 
& {\scriptsize\textcolor{Black}{+0.00}} 81.33 
& {\scriptsize\textcolor{ForestGreen}{+2.43}} 89.69 
& {\scriptsize\textcolor{ForestGreen}{+5.06}} 76.68 
& {\scriptsize\textcolor{ForestGreen}{+3.41}} 13.77 
& {\scriptsize\textcolor{ForestGreen}{+11.83}} 37.53 
& {\scriptsize\textcolor{Red}{-0.21}} 33.68 \\

\midrule
\multirow{3}{*}{\rotatebox[origin=c]{90}{\scriptsize\textbf{SigLIP2}}}
& \full{} & 82.12 & 92.22 & 76.15 & 74.40 & 91.95 & 82.42 & 60.93 & 60.68 & 85.56 \\
& \fg{} 
& {\scriptsize\textcolor{Red}{-4.97}} 
77.15 
& {\scriptsize\textcolor{Red}{-4.34}} 
87.88 
& {\scriptsize\textcolor{Red}{-4.28}} 
71.87 
& {\scriptsize\textcolor{Red}{-17.07}} 
57.33 
& {\scriptsize\textcolor{Red}{-0.99}}
90.96 
& {\scriptsize\textcolor{ForestGreen}{+0.60}} 83.02 
& {\scriptsize\textcolor{ForestGreen}{+3.70}} 64.63 
& {\scriptsize\textcolor{Red}{-1.24}}
59.44 
& {\scriptsize\textcolor{Red}{-3.84}} 81.72 \\
& \fg{}$\oplus$\full{} 
& {\scriptsize\textcolor{ForestGreen}{+0.29}} 82.41 
& {\scriptsize\textcolor{ForestGreen}{+0.15}} 92.37 
& {\scriptsize\textcolor{ForestGreen}{+0.68}} 76.83 
& {\scriptsize\textcolor{Red}{-1.33}} 
73.07 
& {\scriptsize\textcolor{ForestGreen}{+0.63}} 92.58 
& {\scriptsize\textcolor{ForestGreen}{+1.89}} 84.31 
& {\scriptsize\textcolor{ForestGreen}{+4.94}} 65.87 
& {\scriptsize\textcolor{ForestGreen}{+3.32}}
64.00 
& {\scriptsize\textcolor{ForestGreen}{+0.51}} 86.07 \\

\bottomrule
\end{tabular}

\caption{
  Accuracy of ConvNeXT-Tiny, from Timm \cite{rw2019timm},  trained on the  standard (i.e. \full{}) ImageNet-1K data and of SigLIP2-SO400M \cite{tschannen2025siglip} on image classification datasets.
  The evaluation is on 9 dataset, specified in sec \cref{subsec:i1k_datasets},
  with labels a subset of ImageNet-1k  classes. 
  The models classify in all cases into 1k ImageNet classes, even if the test set for a particular dataset does not contain all of them, to keep the task uniform. 
  The script-size numbers indicate the increase/decrease with respect to the baseline, i.e., recognition on the \full{} image. 
  {\bf \fg{}$^+$} denotes results with a model fine-tuned on \fg{} crops (not easily applicable to SigLIP2). It can be observed that finetuning is beneficial and improves both \fg{} and fusion results.
  Hard IN lower performance is due to \fg{} localization problems, see text.
 }
\label{tab:in1k}
\end{table*}

%% file: tables/rw_comp/spaw_rw_comp.tex
\begin{table}[tbh]
\small
\setlength{\tabcolsep}{4pt}

\begin{tabular}{lrlr}
    \toprule
    \full{} & 80.90 & CausIRL \cite{chevalley2022invariant} & {\scriptsize\textcolor{ForestGreen}{+8.42}} 89.32 \\
    \fg{}$^+$ & {\scriptsize\textcolor{ForestGreen}{\textbf{+15.48}}} \textbf{96.38} & MMD-AAE \cite{li2018domain} & {\scriptsize\textcolor{Red}{-2.09}} 78.81 \\
    \fg{}$^+\oplus$ \full{} & {\scriptsize\textcolor{ForestGreen}{+11.01}} \underline{91.91} & Fish \cite{shi2021gradient} & {\scriptsize\textcolor{Red}{-3.39}} 77.51 \\
    \cmidrule(l){1-2}
    ERM \cite{vapnik1991principles} & {\scriptsize\textcolor{Red}{-3.41}} 77.49 & W2D \cite{huang2022two} & {\scriptsize\textcolor{ForestGreen}{+1.04}} 81.94 \\
    GroupDRO \cite{sagawa2019distributionally} & {\scriptsize\textcolor{Red}{-0.32}} 80.58 & JTT \cite{liu2021just} & {\scriptsize\textcolor{ForestGreen}{+9.34}} 90.24 \\
    IRM \cite{arjovsky2019invariant} & {\scriptsize\textcolor{Red}{-5.45}} 75.45 & Mixup \cite{xu2020adversarial} & {\scriptsize\textcolor{ForestGreen}{+7.58}} 88.48 \\
    CORAL \cite{sun2016deep} & {\scriptsize\textcolor{ForestGreen}{+8.76}} 89.66 & Mixup \cite{yao2022improving} & {\scriptsize\textcolor{ForestGreen}{+7.74}} 88.64 \\
    \bottomrule
\end{tabular}

    \caption{Spawrious \cite{lynch2023spawrious}, a dataset with an adversarial \bg{} shift -- comparison to domain generalization methods.
    The \textbf{best} and \underline{second best} results are highlighted.
    All methods 
    use ResNet50. 
    }
    \label{tab:spaw:r50}
\end{table}

%% file: tables/summary.tex
\begin{table}[bth]
\small
\setlength{\tabcolsep}{4pt}
\centering
\begin{tabular}{l r r r}
    \toprule
    \cmidrule(r){1-1} \cmidrule(r){2-4} 
    \textbf{\convnext-tiny} & \textbf{Dogs} & \textbf{Spaw} & \textbf{Fungi}  \\
    \cmidrule(r){2-4}
    \full{} & 72.42 & 37.91 & 47.57   \\
    \fg{}$^+$ & {\scriptsize\textcolor{ForestGreen}{+5.09}} 77.51 & {\scriptsize\textcolor{ForestGreen}{+56.40}} 94.31 & {\scriptsize\textcolor{Red}{-2.61}} 44.96 \\
    \fg{}$^+\oplus$\full{} & {\scriptsize\textcolor{ForestGreen}{+5.77}} 78.19 & {\scriptsize\textcolor{ForestGreen}{+45.04}} 82.95 & {\scriptsize\textcolor{ForestGreen}{+0.60}} 48.17 \\
    \midrule
    
    \textbf{SigLIP2} & & & \textbf{BioCLIP}  \\
    \cmidrule(r){2-4}
    \full{} & 84.37 & 95.33 & 18.58   \\
    \fg{} & {\scriptsize\textcolor{Red}{-0.47}} 83.90 & {\scriptsize\textcolor{ForestGreen}{+1.13}} 96.46 & {\scriptsize\textcolor{Red}{-4.94}} 13.64 \\
    \fg{}$\oplus$\full{} & {\scriptsize\textcolor{ForestGreen}{+0.77}} 85.14 & {\scriptsize\textcolor{ForestGreen}{+1.10}} 96.43 & {\scriptsize\textcolor{ForestGreen}{+0.21}} 18.79 \\

    \bottomrule
\end{tabular}

\caption{
 Recognition accuracy of \fg{}, \bg{}, \full{} and fusion on Stanford Dogs and Spawrious. For FungiTastic, \ac{vlm} results are reported with the domain-specific BioCLIP model.
 }
\label{tab:finegrained}
\end{table}

%% file: tables/rw_comp/guided_crop.tex
\begin{table}[h]
\centering
\small
\setlength{\tabcolsep}{2pt}
\begin{tabular}{l r r r r}
\hline
 & IN-1K & IN-V2 & IN-A & IN-R\\
\cmidrule{2-5}
GC FULL \cite{saranrittichai2024zeroshot} & 58.79 & 51.88 & 29.37 & 65.26 \\
GC \cite{saranrittichai2024zeroshot} & {\scriptsize\textcolor{ForestGreen}{+1.05}} 59.84  & {\scriptsize\textcolor{ForestGreen}{+1.42}} 53.30  & {\scriptsize\textcolor{ForestGreen}{+2.60}} 31.97 & {\scriptsize\textcolor{ForestGreen}{+1.41}} 66.67 \\
\hline
FULL & 59.78 & 52.08 & 26.96 & 63.66 \\
FG~$\oplus$~FULL & {\scriptsize\textcolor{ForestGreen}{+2.50}} 62.28  & {\scriptsize\textcolor{ForestGreen}{+3.47}} 55.55 & {\scriptsize\textcolor{ForestGreen}{+9.59}} 36.55 & {\scriptsize\textcolor{ForestGreen}{+0.78}} 64.44  \\
\hline
\end{tabular}
\caption{Comparison to Guided Cropping (GC) \cite{saranrittichai2024zeroshot} with CLIP-B.
}
\label{tab:rw_comp:guided_crop}
\end{table}

%% file: tables/rw_comp/waterbirds.tex
\begin{table}[h!]
\centering
\small
\setlength{\tabcolsep}{4pt}
\begin{tabular}{l c c c} \toprule \textbf{Method} & \textbf{Additional labels} & \textbf{worst} & \textbf{total} \\ \midrule GroupDRO \cite{sagawa2019distributionally} & \tikzcmark & 89.3 & 94.4 \\ DaC \cite{noohdani2024decompose} & \tikzcmark & 92.6 & 94.9 \\ DaC-C \cite{noohdani2024decompose} & \tikzcmark & 92.3 & 95.3 \\ \midrule ERM \cite{asgari2022masktune} & \tikzxmark & 80.8 & 94.0 \\ MaskTune \cite{asgari2022masktune} & \tikzxmark & 86.4 & 93.0 \\ \midrule \full{} & \tikzxmark & 88.5 & 94.0 \\ \fg{} & \tikzxmark & 92.1 & 96.3 \\ \fg{} $\oplus$ \full{} & \tikzxmark & 91.4 & 95.9 \\ \bottomrule \end{tabular}
\caption{Waterbirds \cite{sagawa2019distributionally} -  comparison to domain generalization methods. Worst-group and total acc (\%). Our method, MaskTune and ERM use the \emph{corrected} dataset from \cite{asgari2022masktune}. All use ResNet50.}
\label{tab:waterbirds}
\end{table}

%% file: sections/conclusion.tex
\section{Conclusion}
\label{sec:conclusion}
This paper introduced a robust context-aware object recognition framework.
By leveraging zero-shot class-agnostic localization, the method enables accurate recognition of objects using their intrinsic features, while still allowing for the controlled and interpretable inclusion of contextual cues.
We demonstrated that this approach improves generalization across domain-shifted benchmarks and maintains or even enhances in-domain performance. The method is simple, non-parametric, and applicable to both supervised models and vision-language models without requiring fine-tuning, while additional benefits can be gained from object-centric model fine-tuning.

Localization was shown to be the main limiting factor,
preventing gains over the baseline on the HardImageNet dataset and with very large supervised models.

%% file: sections/app_embeddings.tex
\section{Training-time localization}
\label{sec:app_loc}
This section provides additional details on \fg{} localization for training/finetuning the object-centric model, \fg{}$^+$.

Recent advances in object detection, such as the OWL and OWLv2 models \cite{minderer2022simple, minderer2024scaling}, have demonstrated that the idea of embedding images and texts into a shared space transfers successfully to detection. Moreover, it allows for replacing text embeddings with suitable image-derived query embeddings, leading to few-shot
detection in cases where linguistic designations are unknown or do not work well, such as unique or specialized classes. See \cref{fig:hard_disk}, \ref{fig:screw}.   

In \cref{sec:class_specific_loc}, we detail on how the prompt embeddings are used for class-aware \fg{} localisation and in \cref{sec:class_specific_loc}, we provide a general algorithm for obtaining effective image-derived query embeddings under minimal assumptions. In particular, this procedure applies (Sec \ref{susbsec:bbox_gen}) to the OWLv2 detector on the ImageNet-1k dataset, as described in \cref{sec:emb_for_imagenet}.  

We assume the detector $ \mathcal{D}' $ encodes each region proposal (bounding box) $ \text{b}$ as a feature vector $ e_{\text{b}} \in \mathbb{R}^N $.
Text prompts are embedded into the same space
via a text encoder, as $ e_{\text{text}} \in \mathbb{R}^N $ . 

\subsection{Class-specific \fg{} localization}
\label{sec:class_specific_loc}
For training an \fg{} model (Sec \ref{sec:imdec}, \ref {susbsec:bbox_gen}) a bounding box associated to the GT label must be identified. 
Given (text or image) queries $q \in \{e_{\text{text}}, e_{\text{img\_q}}\}$ associated to the labels, for each image in the training set a bounding box is selected from a list $ \{ \text{b}_k  \}_{k \in \mathcal{K}} $ (provided by $ \mathcal{D}' $) by choosing the $ k $ which maximizes the cosine similarity $\cos(e_{\text{b}_k}, q )$ 
 
\subsection{Class embeddings for image-conditioned detection}
\label{sec:img_query_embeddings}

As explained above, the step in \ref{sec:class_specific_loc} applies equally to text or image-derived embeddings, since they live in a shared space. While obtaining text embeddings is straightforward, as they are given by a text encoder, defining image-conditioned embeddings is more involved. The present subsection is devoted to this task.

Computing suitable image-derived query embeddings benefits from input images that unambiguously depict a single dominant object representing the target class. To carry out this requirement, define the \emph{objectness ratio} $ \gamma = \frac{s_1}{s_2} $ where $ s_1 $ and $ s_2$ denote the highest and second-highest objectness scores. This ratio serves as a proxy for image suitability: elevated $ \gamma \gg 1 $ indicates strong dominance of a single object. 

If a text encoder and a list of prompts are available, they can be used to filter out images where the top objectness box does not coincide with the box that has the highest text-prompt score.

As shown in ~\cite[Tab. 3]{minderer2022simple} on the detection dataset COCO AP50, few-shot localization (by averaging multiple query embeddings) outperforms one-shot single-query detection. 

Putting all these together we arrive at the steps in Algorithm \ref{alg:class_embedding}.

\subsection{Application to ImageNet-1k}
\label{sec:emb_for_imagenet}

In this section we present the ImageNet-1k class-aware \fg{} localisation by means of the OWLv2 detector \cite{minderer2024scaling}. 

We begin with a list of text prompts available at \cite{imagenet_prompts}, one for each class. These lead to an initial set of bounding boxes that can be used for training.

However, we have noticed that detection by text alone fails on some classes. This is verified by computing IoUs between the generated boxes and GT boxes on a subset of the dataset that has GT bounding box annotations. In \cref{tab:iou_bottom_classes} we list examples of these classes. 

Inspired by \cite[Sec. 4.4]{minderer2022simple}, we have noticed improved detection by replacing text embeddings with image-derived query embeddings, see \cref{tab:iou_bottom_classes}. Example improvements can be seen in \cref{fig:hard_disk}, \ref{fig:screw}. 

The image-derived embeddings are created using the method from \ref{sec:img_query_embeddings}, applying algorithm \ref{alg:class_embedding} (including step 2), with $ k=20 $.

Based on mean GT IoU, image queries outperformed text queries for 651 ImageNet-1k classes, while text queries performed better for 307 classes, with equivalent performance for the remainder.

\input{tables/detection_ious}

\input{figures/box_figures}

%% file: tables/detection_ious.tex
\begin{table}[bth]
\centering
\small
\setlength{\tabcolsep}{4pt}
\begin{tabular}{lcc}
    \toprule
        Class Number and Name        & Text IoU \% & Image IoU \%          \\
    \midrule
    592 - hard disk drive             & 21 & 81           \\
    638 - tights, a type of clothing             & 24 & 71  \\
    677 - metal nail & 24 & 61\\
    616 - knot                & 26 & 44               \\
    783 - screw            & 40 & 78        \\
    \bottomrule
  \end{tabular}
\caption{The 5 ImageNet classes with the most improvement in detection by switching from text embeddings to image-derived embeddings. We record the mean IoU between GT boxes and boxes generated by prompting OWLv2 with: text and image queries.}
\label{tab:iou_bottom_classes}
\end{table}

%% file: figures/box_figures.tex
\begin{figure}
\centering
\includegraphics[width=\linewidth]{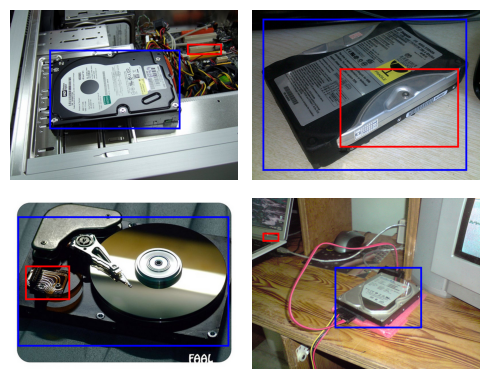}
    \caption{Text (red) vs image query (blue) localisation for the `hard-disk' ImageNet-1k class (592) using OWLv2.}
    \label{fig:hard_disk}
\end{figure}

\begin{figure}
\centering
\includegraphics[width=0.8 \linewidth]{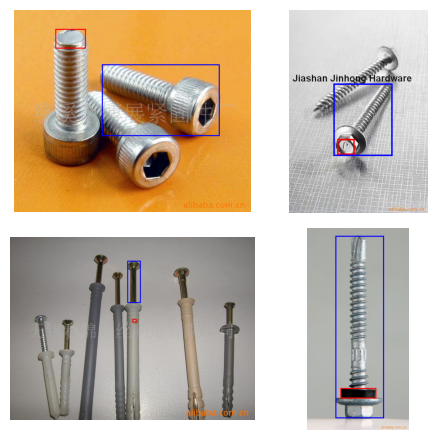}
    \caption{Text (red) vs image query (blue) localisation for the `screw' ImageNet-1k class (783) using OWLv2.}
    \label{fig:screw}
\end{figure}

%% file: sections/app_details.tex
\section{Setup details}
\label{ap:setup:details}
\input{tables/hyperparameters}

\subsection{Evaluation metrics}
\noindent\noindent\textbf{Evaluation metric.} We report the most widely adopted \emph{total accuracy}: 
$ \frac{1}{\text{N}}
{\sum_{i=1}^{\text{N}}
 \left[  y_i = \hat{y}_i \right]
}
$ where $\text{N}$ is the total number of samples and $y_i$, $\hat{y}_i$ are the ground truth and model prediction for image $i$, respectively. 
The metric does not take class imbalance into account. 
The ObjectNet dataset merges some ImageNet classes into one. We treat it as a multilabel dataset, where each image has a set of ground-truth labels $y_i^m \subseteq \mathcal{Y}$, and evaluate with the \emph{ReAL accuracy}:
$\frac{1}{\text{N}} \sum_{i=1}^{\text{N}} \left[ \hat{y}_i \in y_i^m \right]$.

Recognition accuracy is the evaluation metric. For the highly imbalanced FungiTastic, macro-averaged accuracy (mean of per-class accuracies) is reported. 

\subsection{Datasets with ImageNet-1K classes}
While it remains the gold standard for recognition, it contains known labeling flaws~\cite{kisel2024flaws}. To address this, we also evaluate on a ``clean labels'' subset where label corrections from prior work agree~\cite{kisel2024flaws}. The official validation set consists mostly of canonical object-centric images~\cite{recht2019imagenet}, which do not fully represent real-world complexity, where models may exploit shortcuts~\cite{geirhos2020shortcut}. To overcome these limitations, we include the following datasets:

     \noindent\textbf{ImageNet-1K}~\cite{russakovsky2015imagenet}:
    ImageNet-1K is a large-scale image classification dataset. The validation set consists of 50k images across 1,000 object categories. It is widely adopted as the main benchmark for evaluating model performance in visual recognition tasks.
    
     \noindent\textbf{ImageNetV2}~\cite{recht2019imagenet}: Constructed using the original ImageNet collection process to test generalization. We use the `MatchedFrequency' variant, which mirrors the class distribution of the original validation set.

     \noindent\textbf{Hard ImageNet (HIN)}~\cite{moayeri2022hard}: A challenging subset of 15 IN-1K classes with strong \fg{}-\bg{} correlations.
    
     \noindent\textbf{CounterAnimal}~\cite{wang2025sober}: Includes 45 IN-1K animal classes sourced from iNaturalist, with each image labeled by \bg{} rarity (``common'' or ``rare'').
    
     \noindent\textbf{ImageNet-A}~\cite{hendrycks2021natural}: A natural adversarial benchmark composed of 200 ImageNet classes intentionally filtered to induce classification errors.
    
     \noindent\textbf{ObjectNet}~\cite{barbu2019objectnet}: Contains everyday objects under controlled changes in background, viewpoint, and rotation; we evaluate on the 113 classes overlapping with IN-1K.
    
     \noindent\textbf{ImageNet-R}~\cite{hendrycks2021many}: Focuses on 200 ImageNet classes rendered in artistic or abstract forms (e.g., paintings, sculptures), used to assess robustness to distribution shift.

%% file: tables/hyperparameters.tex
\begin{table}[bth]
\centering
\small
\begin{tabular}{lc}
    \toprule
    Hyperparameter        & Value           \\
    \midrule
    Optimizer             & AdamW           \\
    Scheduler             & One-cycle  \\
    Max LR & $5\times10^{-5}$\\
    Epochs                & 20 (\tiny{10\% warmup})               \\
    Batch size            & 256\tiny{$\times$2 GPUs}        \\
    Weight decay          & 0.05            \\
    Layer‐wise decay      & 0.9             \\
    Input size      & 224$\times$224  \\
    MixUp, CutMix       & 0.8, \ 1.0               \\
    Label smoothing       & 0.1             \\
    Random erase p.    & 0.25            \\
    AutoAugment  & {\tiny\texttt{rand-m9-mstd0.5-inc1}} \\ 
    \bottomrule
  \end{tabular}
\caption{Image-Net \fg{}$^+$ fine-tuning hyperparameters. The standard model pretrained on \full{} images is fine-tuned on \fg{} cropped images. Results with the fine-tuned model are denoted as \fg{}$^+$ in the results Tables. Training completes in $ \approx 12 $ hours on two V100 GPUs}
\label{tab:hyperparams}
\end{table}

\begin{table}[bth]
\centering
\small
\begin{tabular}{lc}
    \toprule
    Hyperparameter        & Value           \\
    \midrule
    Optimizer             & AdamW           \\
    Scheduler             & One-cycle  \\
    Max LR & $ 10^{-4}$\\
    \qquad  - FungiTastic \fg{}, Waterbirds & $5\times10^{-5}$   \\

    Epochs:                & 20 (\tiny{10\% warmup})               \\
    \qquad  - Spawrious & 5 (\tiny{10\% warmup})               \\
    \qquad  - FungiTastic \fg{} & 10 (\tiny{10\% warmup})               \\
    Batch size            & 64       \\
    Weight decay          & 0.01            \\
    Input size      & 224$\times$224  \\
    CutMix, Label smoothing: &  - , \  -     \\
    \qquad - FungiTastic   & 0.5, \ 0.1 \\      
    Random erase p.    & 0.25            \\
    AutoAugment  & {\tiny\texttt{rand-m9-mstd0.5-inc1}} \\ 
    Initial timm \cite{rw2019timm} ckpt:  & \\ 
    \qquad - StanfordDogs \full{} and \fg{}    &  {\tiny\texttt{convnextv2\_tiny.fcmae}} \\  
    \qquad - Spawrious \full{} and \fg{}   & {\tiny\texttt{resnet50.a1\_in1k}} \\  
     \qquad - Waterbirds \full{} and \fg{}   & {\tiny\texttt{resnet50.tv\_in1k}} \\  
    \qquad - FungiTastic \full{} &  {\tiny\texttt{convnextv2\_tiny.fcmae\_ft\_in1k}} \\ 
    \qquad - FungiTastic \fg{} &  {\tiny\texttt{ FungiTastic \full{} trained}} \\ 
    \bottomrule
  \end{tabular}
\caption{Hyperparameters for training fine-grained datasets: FungiTastic, Spawrious, Stanford Dogs and Waterbirds. The default value for all datasets is given, unless otherwise specified. }
\label{tab:hyperparams2}. 
\end{table}

%% file: sections/app_exps.tex
\section{Experiments}
\label{ap:ablations}
\input{tables/fungi_meta}
\input{figures/tradeoff}

\input{tables/convnext_weight_ablation}

\input{tables/siglip2_in1k_ablations}
\input{tables/convnext_in1k_gt_prompts}
\input{tables/in1k_eva}
\input{tables/rw_comp/dfr_shape_comp}

\noindent\textbf{Multi-Object resolution strategies.} Many images in general classification datasets are complex, multi-object scenes. For \fg{} and \full{} fusion, a single object needs to be selected as the target object most likely to correspond to the ground-truth object. \cref{tab:siglip2_in1k_abl} compares different multiobject resolution strategies. The proposed approach from the main paper, based on objectness-weighted maximum confidence, is denotes as \fg{}$_\text{OWMC}$. We compare it to 3 alternative strategies. \fg{}$_\text{HS}$  selects the object with the highest objectness score, \fg{}$_\text{U}$ creates a single \fg{} region as the union of all the candidate object regions, and \fg{}$_\text{MA}$ selects the object with the maximum axis-aligned crop area.
The results show that while the simpler alternative strategies may be enough, even preferable for certain dataset, the proposed method performance is stable across all the test sets.

We can observe that on the OOD datasets, the strategies resulting in larger image areas, \fg{}$_\text{U}$ and \fg{}$_\text{HS}$,  tend to underperform the proposed \fg{}$_\text{OWMC}$ and the \fg{}$_\text{HS}$.
On ID data, \fg{}$_\text{U}$ and \fg{}$_\text{MA}$  outperform \fg{}$_\text{OWMC}$ and \fg{}$_\text{HS}$, being allowed to rely on more contextual information.

While the strategies perform similarly when fused with \full{} on ID datasets, on OOD data, the benefits of the more dominant-object centric strategies show.

\noindent\textbf{Oracle (GT) prompt localization for evaluation.}
While GT bounding boxes are not available for most datasets, the same scheme for bounding box generation from \cref{susbsec:bbox_gen}  can be applied to the test set, obtaining ``GT prompt" object crops. 
Evaluating the model (fine-tuned on image crops) on these oracle-prompt test crops then provides an upper bound of what can be obtained by our method, at least under idealized conditions (e.g., we approximate the scenario if detection were perfect).
The results are reported in \cref{tab:gtprompts}.

\noindent\textbf{BG model with FungiTastic metadata.}
In the main paper, the contextual \bg{} is always modelled as part of the \full{} image.
This experiment explores an alternative approach to \bg{} modelling based on tabular metadata.
The FungiTastic dataset comes with various additional data, we pick \textit{habitat}, \textit{substrate} and \textit{month}, which are highly related to the \bg{} appearance. Inspired by the metadata prior model of \cite{Picek_2022_WACV,berg2014birdsnap,picek2024fungitastic}, the method precomputes a prior probability of each (class, metadata) value combination and re-weights the classifier predictions based on the metadata. The metadata-prior model assumes the appearance of the image is independent of the metadata, which is not true when the image \bg{} is included (such as in the case of \full{}). Combining with \fg{} makes the method more principled.
The localization in this experiment was performed by prompting the detector with the text `a mushroom'.

Results in Table \ref{tab:fungi_meta} show that all metadata kinds improve the performance of both \fg{}$^+$ and \full{} \convnext-Base models. The habitat helps the most, adding 3.8 \% to the 43.5 \% baseline of \full{} and 4.2 \% to the 44 \% baseline of \fg{}. For habitat and month, the improvements from metadata fusion are greater for the \fg{} than for the \full{}, even though the \fg{} already performs better than \full{}. We hypothesize this can be due to the suppression of \bg{} influence in \fg{}$^+$, leading to better \fg{}-\bg{} decoupling, as assumed by the metadata-prior model.

\noindent\textbf{Robust/context-aware recognition trade-off.}
The trade-off between robust (\fg{}) and context-aware (\full{}) recognition is visualized in \cref{fig:app:exps:tradeoff}. We summarize the model ID and OOD performance by averaging its performance on the corresponding ImageNet test sets. 

\noindent\textbf{Comparison to Deep Feature Reweighting (DFR).} A comparison of \methodabb{} to DFR \cite{kirichenko2023last} (shape bias experiment, Table 3 in \cite{kirichenko2023last}) on ImageNet-1K and ImageNet-R is provided in \cref{tab:dfr_shape_comp}. While our torchvision model's initial performance is slightly lower (possibly due to different input image transformation, or because the weights of the model have been updated), \methodabb{} significantly outperforms DFR on ImageNet-1K (where DFR underperforms the baseline) while remaining competitive on ImageNet-R. Even the baseline \full{} with the Timm weights outperforms all other methods.

\subsection{Alternative approaches}
\noindent\textbf{Context-aware models}.
We propose a method to disentangle object-centric and context-aware representations, where context is defined as the \full{} image (comprising both \fg{} and \bg{}). We explored different ways to extract context-aware representations by masking out the \fg{}, using either bounding boxes or segmentation masks. Bounding box masking significantly underperformed, while the more accurate but computationally expensive mask-based removal (requiring an external model like Segment Anything \cite{kirillov2023segment}) showed notable gains only on the Spawrious datasets—an extreme case where segmentation is trivial. Although separating \bg{} from \full{} could be beneficial (especially when \fg{} alone suffices for recognition, rendering \full{} uninformative for context-aware recognition), our simple experiments did not demonstrate such gains.

\noindent\textbf{Fusion}. We also explored more complex fusion methods, including temperature-scaled logit averaging and learned fusion (\ie fully-connected layers). While learned fusion can offer benefits, particularly when training and evaluation data share similar distributions, its effectiveness is highly data-dependent and does not generalize well across scenarios.

\noindent\textbf{Localizers}. Preliminary experiments with alternative localization models were conducted, namely GroundingDINO \cite{liu2023grounding} (an open-vocabulary detector) and CutLER \cite{wang2023cut} (a class-agnostic, self-supervised segmentation model). While GroundingDINO performs well on fine-grained datasets when the object of the prompted class is always in the image, it tends to predict objects with high confidence regardless of their presence, limiting its effectiveness for general datasets like ImageNet. CutLER significantly underperformed compared to OWLv2.

%% file: tables/fungi_meta.tex
\begin{table}[bth]
\setlength{\tabcolsep}{4pt}
    \centering
    \begin{tabular}{lrrrr}
    \toprule
     {\sc bg:}
     & \multicolumn{1}{c}{\textbf{}} & \multicolumn{1}{c}{\textbf{+habitat}} & \multicolumn{1}{c}{\textbf{+substrate}} & \multicolumn{1}{c}{\textbf{+month}} \\
    \midrule
    \full{} & 43.50 & 47.26 {\color{darkgreen}\scriptsize +3.77} & 45.42 {\color{darkgreen}\scriptsize +1.92} & 45.19 {\color{darkgreen}\scriptsize +1.70} \\
    \fg{}$^+$ & 44.00 & 48.22 {\color{darkgreen}\scriptsize +4.22} & 45.77 {\color{darkgreen}\scriptsize +1.77} & 45.80 {\color{darkgreen}\scriptsize +1.81} \\
    \bottomrule
    \end{tabular}
    \caption{Mean class accuracy of fusion models with \bg{} representation \cite{berg2014birdsnap,Picek_2022_WACV} based on tabular metadata (habitat, substrate, month) on the FungiTatsic dataset. The increment over image-only performance is also reported. The results are averaged across 5 runs with different random seeds.}
    \label{tab:fungi_meta}
\end{table}

%% file: figures/tradeoff.tex
\begin{figure}[bt]
\centering
\includegraphics[width=0.98\linewidth]{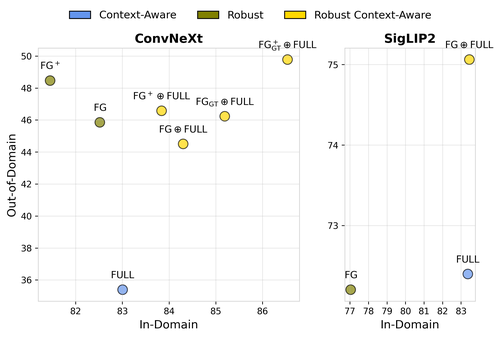}
\caption{The trade-off between robust (\fg{}) and context-aware (\full{}) recognition, visualized as averaged performance across in-domain (ID) and out-of-domain (OOD) datasets. }
\label{fig:app:exps:tradeoff}
\end{figure}

%% file: tables/convnext_weight_ablation.tex
\begin{table*}[ht]
\small
\setlength{\tabcolsep}{5pt}
\centering
\begin{tabular}{c l ccccc ccccc}
\toprule
\cmidrule{3-11}
& & \multicolumn{5}{c}{\textbf{BG important or no domain shift}} & \multicolumn{5}{c}{\textbf{BG uninformative or adversarial}} \\
\cmidrule(l){3-7}
\cmidrule(l){8-12}
& method & IN-1K:Val & IN-1K:Clean & IN-V2 & Hard IN & Animal-C & Animal-R & IN-A & Obj-N & IN-R \\
\midrule

\multirow{8}{*}{\rotatebox[origin=c]{90}{{\scriptsize\textbf{ConvNeXT}}}}
& \fg{} weighted & 82.23 & 92.88 & 71.93 & 76.40 & 89.15 & 77.64 & 29.99 & 39.25 & 36.53 \\
& \fg{} no weight & 81.80 & 92.72 & 71.52 & 77.87 & 89.21 & 77.34 & 27.48 & 36.14 & 35.56 \\
\cmidrule{2-11}
& \fg{}$^+$ weighted & 81.48 & 92.15 & 71.55 & 72.27 & 89.82 & 78.74 & 37.95 & 39.32 & 37.84 \\
& \fg{}$^+$ no weight & 80.32 & 91.36 & 69.89 & 72.40 & 89.23 & 77.47 & 32.84 & 34.11 & 36.08 \\
\cmidrule{2-11}
& \fg{}$\oplus$\full{} weighted & 83.38 & 93.86 & 72.98 & 81.33 & 89.96 & 77.69 & 25.64 & 37.86 & 36.83 \\
& \fg{}$\oplus$\full{} no weight & 82.35 & 93.23 & 71.82 & 79.07 & 89.86 & 77.62 & 25.07 & 35.53 & 36.03 \\
\cmidrule{2-11}
& \fg{}$^+\oplus$\full{} weighted & 83.32 & 93.78 & 72.86 & 79.20 & 90.04 & 77.94 & 31.81 & 38.38 & 38.17 \\
& \fg{}$^+\oplus$\full{} no weight & 81.68 & 92.70 & 71.00 & 76.93 & 89.55 & 77.35 & 29.45 & 34.09 & 36.89 \\
\bottomrule
\end{tabular}
\caption{The effect of removing the weights from the \fg{} prediction \eqref{eq:weighted_confidence}.
Accuracy of ConvNeXT-Tiny trained on \full{} ImageNet data from Timm \cite{rw2019timm} (\fg{}$^+$ denotes results with model finetuned on \fg{}) on datasets with ImageNet (IN) classes. All models are evaluated against all of the 1k ImageNet classes, even if the test set does not contain all of them.
}
\label{tab:noweights}
\end{table*}

%% file: tables/siglip2_in1k_ablations.tex
\begin{table*}[ht]
\small
\setlength{\tabcolsep}{3pt}
\centering
\begin{tabular}{c l rrrrr rrrrr}
\toprule
 & method $\downarrow$  & \multicolumn{5}{c}{\textbf{BG informative, no domain shift}} & \multicolumn{5}{c}{\textbf{BG uninformative or adversarial}} \\
\cmidrule(l){3-7}
\cmidrule(l){8-12}

& datasets $\rightarrow$ 
& 
\multicolumn{1}{c}{  IN-1K:Val} & 
\multicolumn{1}{c}{  IN-1K:Clean} & 
\multicolumn{1}{c}{  IN-V2} & 
\multicolumn{1}{c}{  Hard IN} & 
\multicolumn{1}{c}{  Animal-C} & 
\multicolumn{1}{c}{Animal-R} & 
\multicolumn{1}{c}{IN-A} & 
\multicolumn{1}{c}{  Object-Net} & 
\multicolumn{1}{c}{  IN-R} \\
\midrule
\multirow{9}{*}{\rotatebox[origin=c]{90}{\scriptsize\textbf{SigLIP2}}}
& \full{} & 82.12 & 92.22 & 76.15 & 74.40 & 91.95 & 82.42 & 60.93 & 60.68 & 85.56 \\

\cmidrule{2-11}

& \fg{}$_\text{OWMC}$ 
&  77.15 
&  87.88 
&  71.87 
&  57.33 
&  90.96 
&  83.02 
&  64.63 
& 59.44 
& 81.72 \\

& \fg{}$_\text{HS}$
& 74.99 & 85.52 & 69.45 & 49.47 & 89.81 & 82.99 & 60.55 & 63.68 & 82.16 \\

& \fg{}$_\text{U}$  
& 79.15 & 88.99 & 73.08 & 68.67 & 90.09 & 82.60 & 60.51 & 60.33 & 82.22 \\

& \fg{}$_\text{MA}$ 
& 78.06 & 88.06 & 71.78 & 64.40 & 88.85 & 81.20 & 54.85 & 57.45 & 82.29 \\

\cmidrule{2-11}
& \fg{}$_\text{OWMC} \oplus$\full{} 
&  82.41 
&  92.37 
&  76.83 
&  73.07 
&  92.58 
&  84.31 
&  65.87 
& 64.00 
& 86.07 \\

& \fg{}$_\text{HS} \oplus$\full{}
& 82.18 & 92.19 & 76.43 & 71.47 & 92.45 & 84.32 & 66.45 & 65.76 & 86.19 \\

& \fg{}$_\text{U} \oplus$\full{}  
& 82.47 & 92.41 & 76.50 & 73.60 & 92.32 & 83.51 & 63.97 & 62.15 & 85.93 \\

& \fg{}$_\text{MA} \oplus$\full{} 
&  82.27 & 92.28 & 76.50 & 73.20 & 92.05 & 83.40 & 63.15 & 61.61 & 85.89 \\

\bottomrule
\end{tabular}

\caption{
Different \fg{} object selection criteria.
  Accuracy of ConvNeXT-Tiny, from Timm \cite{rw2019timm},  trained on the  standard (i.e. \full{}) ImageNet-1K data and of SigLIP2-SO400M \cite{tschannen2025siglip} on image classification datasets. \fg{}$_\text{OMWC}$, \fg{}$_\text{HS}$, \fg{}$_\text{U}$ and \fg{}$_\text{MA}$ correspond to the objectness-weighted maximum confidence prediction from the main paper (\cref{eq:weighted_confidence}, \eqref{eq:fusion_def}), highest-score, union of all boxes and maximum-area multi-object resolution strategies, respectively.
 }
\label{tab:siglip2_in1k_abl}
\end{table*}

%% file: tables/convnext_in1k_gt_prompts.tex
\begin{table*}[ht]
\small
\setlength{\tabcolsep}{5pt}
\centering
\begin{tabular}{c l ccccc ccccc}
\toprule
\cmidrule{3-11}
& & \multicolumn{5}{c}{\textbf{BG important or no domain shift}} & \multicolumn{5}{c}{\textbf{BG uninformative or adversarial}} \\
\cmidrule(l){3-7}
\cmidrule(l){8-12}
& method & IN-1K:Val & IN-1K:Clean & IN-V2 & Hard IN & Animal-C & Animal-R & IN-A & Obj-N & IN-R \\
\midrule

\multirow{9}{*}{\rotatebox[origin=c]{90}{{\scriptsize\textbf{ConvNeXT}}}}
& \full{} & 82.35 & 93.12 & 70.97 & 81.33 & 87.26 & 71.62 & 10.36 & 25.70 & 33.89 \\
\cmidrule{2-11}
& \fg{}$^+$ & 81.48 & 92.15 & 71.55 & 72.27 & 89.82 & 78.74 & 37.95 & 39.32 & 37.84 \\
& \fg{}$^+_{\text{GT}}$ & 85.95 & 93.61 & 77.45 & 80.93 & 90.09 & 78.50 & 49.44 & 47.61 & 40.60 \\
\cmidrule{2-11}
& \fg{} & 82.23 & 92.88 & 71.93 & 76.40 & 89.15 & 77.64 & 29.99 & 39.25 & 36.53 \\
& \fg{}$_{\text{GT}}$ & 83.68 & 92.82 & 74.24 & 66.27 & 89.47 & 77.73 & 36.24 & 44.83 & 37.36 \\

\cmidrule{2-11}
& \fg{}$^+\oplus$\full{} & 83.32 & 93.78 & 72.86 & 79.20 & 90.04 & 77.94 & 31.81 & 38.38 & 38.17 \\
& \fg{}$^+_{\text{GT}}$ $\oplus$ \full{} & 86.08 & 94.31 & 76.67 & 85.47 & 90.15 & 77.79 & 37.72 & 43.76 & 39.84 \\
\cmidrule{2-11}
& \fg{}$\oplus$\full{} & 83.38 & 93.86 & 72.98 & 81.33 & 89.96 & 77.69 & 25.64 & 37.86 & 36.83 \\
& \fg{}$_{\text{GT}} \oplus$ \full{} & 84.69 & 94.16 & 74.79 & 82.13 & 90.19 & 77.62 & 28.51 & 41.28 & 37.52 \\
\bottomrule
\end{tabular}
\caption{Accuracy of ConvNeXT-Tiny model  from Timm \cite{rw2019timm} evaluated on GT prompts crops. \fg{} uses a model pre-trained on \full{} ImageNet data, \fg{}$^+$ denotes results with model finetuned on \fg{} crops. 
All models are evaluated against all of the 1k ImageNet classes.}
\label{tab:gtprompts}
\end{table*}

%% file: tables/in1k_eva.tex
\begin{table*}[ht]
\small
\setlength{\tabcolsep}{5pt}
\centering
\begin{tabular}{c l ccccc ccccc}
\toprule
\cmidrule{3-11}
& & \multicolumn{5}{c}{\textbf{BG important or no domain shift}} & \multicolumn{5}{c}{\textbf{BG uninformative or adversarial}} \\
\cmidrule(l){3-7}
\cmidrule(l){8-12}
& method & IN-1K:Val & IN-1K:Clean & IN-V2 & Hard IN & Animal-C & Animal-R & IN-A & Obj-N & IN-R \\
\midrule

\multirow{5}{*}{\rotatebox[origin=c]{90}{{\scriptsize\textbf{Eva}}}}
& \full{}          & 90.05 & 96.02 & 82.59 & 90.00 & 94.09 & 85.88 & 67.07 & 57.49 & 73.21 \\
& \fg{}            & 88.56 & 95.36 & 80.95 & 86.13 & 91.68 & 83.68 & 67.32 & 61.87 & 72.16 \\
& \fg{}$\oplus$\full{}      & 89.29 & 95.81 & 81.86 & 87.33 & 93.88 & 85.96 & 69.49 & 61.59 & 72.70 \\
& \fg{}$_{\text{GT}}$     & 89.51 & 95.39 & 82.65 & 79.47 & 92.20 & 83.90 & 71.85 & 68.97 & 73.63 \\
& \fg{}$_{\text{GT}}$ $\oplus$ \full{} & 90.34 & 96.03 & 83.67 & 89.07 & 94.17 & 86.08 & 73.15 & 66.64 & 74.16 \\

\bottomrule
\end{tabular}

\caption{
  Accuracy of pretrained $\text{eva02\_large\_patch14\_448.mim\_m38m'\_ft\_in22k\_in1k} $ from Timm on datasets with ImageNet (IN) classes. All models are evaluated against all of the 1k ImageNet classes, even if the test set does not contain all of them. While performance improves on some OOD datasets, in-domain gains are not achieved in the zero-shot setup. The ablation with ground truth prompt crops shows that localization is the limiting factor and as it improves, the benefits of \methodabb{} will be stronger.
}
\label{tab:methods_vs_datasets_eva}
\end{table*}

%% file: tables/rw_comp/dfr_shape_comp.tex
\begin{table}[h!]
\centering
\small
\setlength{\tabcolsep}{2pt}
\begin{tabular}{l c c c c}
\toprule
\textbf{Method} & \textbf{Zero-Shot} & \textbf{Train. Data}  & \multicolumn{2}{c}{\textbf{Top-1 Acc (\%)}} \\
\cmidrule(lr){4-5}
& & & IN & IN-R  \\
\midrule
\textbf{DFR} \cite{kirichenko2023last} & \\
\full{} &  & IN & 76.0 & 23.8 \\
DFR & False & SIN  & 65.1 & 24.6  \\
DFR & False & IN+SIN & 74.5 & 27.2  \\
\midrule
\textbf{Shape} \cite{geirhos2018imagenet} & \\
\full{}  & False & IN+SIN & 76.8 & 25.6 \\
\midrule
\textbf{RCOR (our)} & \\
\full{} \cite{torchvision2016} &  & IN & 74.57 & 23.47  \\
\full{} $\oplus$ \fg{} & True & IN & 77.51 & 26.57 \\
\cmidrule(lr){2-5}
\full{} \cite{rw2019timm} &  & IN & 79.29 & 27.79  \\
\full{} $\oplus$ \fg{} & True & IN & 81.15 & 30.62 \\
\bottomrule
\end{tabular}
\caption{\textbf{Shape bias experiment from Last Layer Re-Training is Sufficient for Robustness to Spurious Correlations.} Shape bias and accuracy on ImageNet validation set variations for ResNet-50 trained on different datasets (\cite{geirhos2018imagenet}) and DFR with an ImageNet-trained ResNet-50 as a feature extractor (\cite{kirichenko2023last}) compared to \methodabb{} (bottom).
We report results with two different Resnet50 checkpoints, from torchvision \cite{torchvision2016} (top) and from Timm \cite{rw2019timm}.
}
\label{tab:dfr_shape_comp}
\end{table}

%% file: main.bbl
\begin{thebibliography}{81}
\providecommand{\natexlab}[1]{#1}
\providecommand{\url}[1]{\texttt{#1}}
\expandafter\ifx\csname urlstyle\endcsname\relax
  \providecommand{\doi}[1]{doi: #1}\else
  \providecommand{\doi}{doi: \begingroup \urlstyle{rm}\Url}\fi

\bibitem[dog()]{dogs_leaderboard}
Stanford dogs leaderboard.
\newblock \url{https://paperswithcode.com/sota/fine-grained-image-classification-on-stanford-1}.

\bibitem[Acharya et~al.(2022)Acharya, Roy, Koneripalli, Jha, Kanan, and Divakaran]{acharya2022detecting}
Manoj Acharya, Anirban Roy, Kaushik Koneripalli, Susmit Jha, Christopher Kanan, and Ajay Divakaran.
\newblock Detecting out-of-context objects using contextual cues.
\newblock \emph{arXiv preprint arXiv:2202.05930}, 2022.

\bibitem[Aniraj et~al.(2023)Aniraj, Dantas, Ienco, and Marcos]{aniraj2023masking}
Ananthu Aniraj, Cassio~F Dantas, Dino Ienco, and Diego Marcos.
\newblock Masking strategies for background bias removal in computer vision models.
\newblock In \emph{Proceedings of the IEEE/CVF International Conference on Computer Vision}, pages 4397--4405, 2023.

\bibitem[Arjovsky et~al.(2019)Arjovsky, Bottou, Gulrajani, and Lopez-Paz]{arjovsky2019invariant}
Martin Arjovsky, L{\'e}on Bottou, Ishaan Gulrajani, and David Lopez-Paz.
\newblock Invariant risk minimization.
\newblock \emph{arXiv preprint arXiv:1907.02893}, 2019.

\bibitem[Asgari et~al.(2022)Asgari, Khani, Khani, Gholami, Tran, Mahdavi~Amiri, and Hamarneh]{asgari2022masktune}
Saeid Asgari, Aliasghar Khani, Fereshte Khani, Ali Gholami, Linh Tran, Ali Mahdavi~Amiri, and Ghassan Hamarneh.
\newblock Masktune: Mitigating spurious correlations by forcing to explore.
\newblock \emph{Advances in Neural Information Processing Systems}, 35:\penalty0 23284--23296, 2022.

\bibitem[Barbu et~al.(2019)Barbu, Mayo, Alverio, Luo, Wang, Gutfreund, Tenenbaum, and Katz]{barbu2019objectnet}
Andrei Barbu, David Mayo, Julian Alverio, William Luo, Christopher Wang, Dan Gutfreund, Josh Tenenbaum, and Boris Katz.
\newblock Objectnet: A large-scale bias-controlled dataset for pushing the limits of object recognition models.
\newblock \emph{Advances in neural information processing systems}, 32, 2019.

\bibitem[Beery et~al.(2018)Beery, Van~Horn, and Perona]{beery2018recognition}
Sara Beery, Grant Van~Horn, and Pietro Perona.
\newblock Recognition in terra incognita.
\newblock In \emph{Proceedings of the European conference on computer vision (ECCV)}, pages 456--473, 2018.

\bibitem[Berg et~al.(2014)Berg, Liu, Woo~Lee, Alexander, Jacobs, and Belhumeur]{berg2014birdsnap}
Thomas Berg, Jiongxin Liu, Seung Woo~Lee, Michelle~L Alexander, David~W Jacobs, and Peter~N Belhumeur.
\newblock Birdsnap: Large-scale fine-grained visual categorization of birds.
\newblock In \emph{Proceedings of the IEEE conference on computer vision and pattern recognition}, pages 2011--2018, 2014.

\bibitem[Bhatt et~al.(2024)Bhatt, Das, Sigal, and N~Balasubramanian]{bhatt2024mitigating}
Gaurav Bhatt, Deepayan Das, Leonid Sigal, and Vineeth N~Balasubramanian.
\newblock Mitigating the effect of incidental correlations on part-based learning.
\newblock \emph{Advances in Neural Information Processing Systems}, 36, 2024.

\bibitem[Carter()]{imagenet_prompts}
Jimmy Carter.
\newblock Imagenet1k prompts.
\newblock \url{https://huggingface.co/jimmycarter/imagenet1k-clip-big-g-embeds/blob/main/imagenet_zeroshot_data.py}.

\bibitem[Chen et~al.(2023)Chen, Huang, Zhou, Bian, Han, and Cheng]{chen2023understanding}
Yongqiang Chen, Wei Huang, Kaiwen Zhou, Yatao Bian, Bo Han, and James Cheng.
\newblock Understanding and improving feature learning for out-of-distribution generalization.
\newblock \emph{Advances in Neural Information Processing Systems}, 36:\penalty0 68221--68275, 2023.

\bibitem[Cheng et~al.(2022)Cheng, Misra, Schwing, Kirillov, and Girdhar]{cheng2022masked}
Bowen Cheng, Ishan Misra, Alexander~G Schwing, Alexander Kirillov, and Rohit Girdhar.
\newblock Masked-attention mask transformer for universal image segmentation.
\newblock In \emph{Proceedings of the IEEE/CVF conference on computer vision and pattern recognition}, pages 1290--1299, 2022.

\bibitem[Cheng et~al.(2024)Cheng, Oh, Price, Lee, and Schwing]{cheng2024putting}
Ho~Kei Cheng, Seoung~Wug Oh, Brian Price, Joon-Young Lee, and Alexander Schwing.
\newblock Putting the object back into video object segmentation.
\newblock In \emph{Proceedings of the IEEE/CVF Conference on Computer Vision and Pattern Recognition}, pages 3151--3161, 2024.

\bibitem[Chevalley et~al.(2022)Chevalley, Bunne, Krause, and Bauer]{chevalley2022invariant}
Mathieu Chevalley, Charlotte Bunne, Andreas Krause, and Stefan Bauer.
\newblock Invariant causal mechanisms through distribution matching.
\newblock \emph{arXiv preprint arXiv:2206.11646}, 2022.

\bibitem[Chou et~al.(2023)Chou, Kao, and Lin]{chou2023fine}
Po-Yung Chou, Yu-Yung Kao, and Cheng-Hung Lin.
\newblock Fine-grained visual classification with high-temperature refinement and background suppression.
\newblock \emph{arXiv preprint arXiv:2303.06442}, 2023.

\bibitem[Chun and Jiang(1998)]{chun1998contextual}
Marvin~M Chun and Yuhong Jiang.
\newblock Contextual cueing: Implicit learning and memory of visual context guides spatial attention.
\newblock \emph{Cognitive psychology}, 36\penalty0 (1):\penalty0 28--71, 1998.

\bibitem[Deng et~al.(2009)Deng, Dong, Socher, Li, Li, and Fei-Fei]{deng2009imagenet}
Jia Deng, Wei Dong, Richard Socher, Li-Jia Li, Kai Li, and Li Fei-Fei.
\newblock Imagenet: A large-scale hierarchical image database.
\newblock In \emph{2009 IEEE conference on computer vision and pattern recognition}, pages 248--255. Ieee, 2009.

\bibitem[Divvala et~al.(2009)Divvala, Hoiem, Hays, Efros, and Hebert]{divvala2009empirical}
Santosh~K Divvala, Derek Hoiem, James~H Hays, Alexei~A Efros, and Martial Hebert.
\newblock An empirical study of context in object detection.
\newblock In \emph{2009 IEEE Conference on computer vision and Pattern Recognition}, pages 1271--1278. IEEE, 2009.

\bibitem[Fan et~al.(2021)Fan, Fan, Fu, Tang, Shao, and Tai]{fan2021group}
Qi Fan, Deng-Ping Fan, Huazhu Fu, Chi-Keung Tang, Ling Shao, and Yu-Wing Tai.
\newblock Group collaborative learning for co-salient object detection.
\newblock In \emph{Proceedings of the IEEE/CVF Conference on Computer Vision and Pattern Recognition}, pages 12288--12298, 2021.

\bibitem[Geirhos et~al.(2018)Geirhos, Rubisch, Michaelis, Bethge, Wichmann, and Brendel]{geirhos2018imagenet}
Robert Geirhos, Patricia Rubisch, Claudio Michaelis, Matthias Bethge, Felix~A Wichmann, and Wieland Brendel.
\newblock Imagenet-trained cnns are biased towards texture; increasing shape bias improves accuracy and robustness.
\newblock In \emph{International conference on learning representations}, 2018.

\bibitem[Geirhos et~al.(2020)Geirhos, Jacobsen, Michaelis, Zemel, Brendel, Bethge, and Wichmann]{geirhos2020shortcut}
Robert Geirhos, J{\"o}rn-Henrik Jacobsen, Claudio Michaelis, Richard Zemel, Wieland Brendel, Matthias Bethge, and Felix~A Wichmann.
\newblock Shortcut learning in deep neural networks.
\newblock \emph{Nature Machine Intelligence}, 2\penalty0 (11):\penalty0 665--673, 2020.

\bibitem[Ghosh et~al.(2024)Ghosh, Evuru, Kumar, Tyagi, Sakshi, Chowdhury, and Manocha]{ghosh2024aspire}
Sreyan Ghosh, Chandra Kiran~Reddy Evuru, Sonal Kumar, Utkarsh Tyagi, S Sakshi, Sanjoy Chowdhury, and Dinesh Manocha.
\newblock Aspire: Language-guided data augmentation for improving robustness against spurious correlations.
\newblock In \emph{Findings of the Association for Computational Linguistics ACL 2024}, pages 386--406, 2024.

\bibitem[Henderson and Hollingworth(1999)]{henderson1999high}
John~M Henderson and Andrew Hollingworth.
\newblock High-level scene perception.
\newblock \emph{Annual review of psychology}, 50\penalty0 (1):\penalty0 243--271, 1999.

\bibitem[Hendrycks et~al.(2021{\natexlab{a}})Hendrycks, Basart, Mu, Kadavath, Wang, Dorundo, Desai, Zhu, Parajuli, Guo, Song, Steinhardt, and Gilmer]{hendrycks2021many}
Dan Hendrycks, Steven Basart, Norman Mu, Saurav Kadavath, Frank Wang, Evan Dorundo, Rahul Desai, Tyler Zhu, Samyak Parajuli, Mike Guo, Dawn Song, Jacob Steinhardt, and Justin Gilmer.
\newblock The many faces of robustness: A critical analysis of out-of-distribution generalization.
\newblock \emph{ICCV}, 2021{\natexlab{a}}.

\bibitem[Hendrycks et~al.(2021{\natexlab{b}})Hendrycks, Zhao, Basart, Steinhardt, and Song]{hendrycks2021natural}
Dan Hendrycks, Kevin Zhao, Steven Basart, Jacob Steinhardt, and Dawn Song.
\newblock Natural adversarial examples.
\newblock In \emph{Proceedings of the IEEE/CVF conference on computer vision and pattern recognition}, pages 15262--15271, 2021{\natexlab{b}}.

\bibitem[Huang et~al.(2022)Huang, Wang, Huang, Lee, and Xing]{huang2022two}
Zeyi Huang, Haohan Wang, Dong Huang, Yong~Jae Lee, and Eric~P Xing.
\newblock The two dimensions of worst-case training and their integrated effect for out-of-domain generalization.
\newblock In \emph{Proceedings of the IEEE/CVF Conference on Computer Vision and Pattern Recognition}, pages 9631--9641, 2022.

\bibitem[Izmailov et~al.(2022)Izmailov, Kirichenko, Gruver, and Wilson]{izmailov2022feature}
Pavel Izmailov, Polina Kirichenko, Nate Gruver, and Andrew~G Wilson.
\newblock On feature learning in the presence of spurious correlations.
\newblock \emph{Advances in Neural Information Processing Systems}, 35:\penalty0 38516--38532, 2022.

\bibitem[Ke et~al.(2024)Ke, Ye, Danelljan, Tai, Tang, Yu, et~al.]{ke2024segment}
Lei Ke, Mingqiao Ye, Martin Danelljan, Yu-Wing Tai, Chi-Keung Tang, Fisher Yu, et~al.
\newblock Segment anything in high quality.
\newblock \emph{Advances in Neural Information Processing Systems}, 36, 2024.

\bibitem[Khosla et~al.(2011)Khosla, Jayadevaprakash, Yao, and Fei-Fei]{KhoslaYaoJayadevaprakashFeiFei_FGVC2011}
Aditya Khosla, Nityananda Jayadevaprakash, Bangpeng Yao, and Li Fei-Fei.
\newblock Novel dataset for fine-grained image categorization.
\newblock In \emph{First Workshop on Fine-Grained Visual Categorization, IEEE Conference on Computer Vision and Pattern Recognition}, Colorado Springs, CO, 2011.

\bibitem[Kirichenko et~al.(2023)Kirichenko, Izmailov, and Wilson]{kirichenko2023last}
Polina Kirichenko, Pavel Izmailov, and Andrew~Gordon Wilson.
\newblock Last layer re-training is sufficient for robustness to spurious correlations.
\newblock In \emph{The Eleventh International Conference on Learning Representations}, 2023.

\bibitem[Kirillov et~al.(2023)Kirillov, Mintun, Ravi, Mao, Rolland, Gustafson, Xiao, Whitehead, Berg, Lo, et~al.]{kirillov2023segment}
Alexander Kirillov, Eric Mintun, Nikhila Ravi, Hanzi Mao, Chloe Rolland, Laura Gustafson, Tete Xiao, Spencer Whitehead, Alexander~C Berg, Wan-Yen Lo, et~al.
\newblock Segment anything.
\newblock In \emph{Proceedings of the IEEE/CVF International Conference on Computer Vision}, pages 4015--4026, 2023.

\bibitem[Kisel et~al.(2024)Kisel, Volkov, Hanzelkova, Janouskova, and Matas]{kisel2024flaws}
Nikita Kisel, Illia Volkov, Katerina Hanzelkova, Klara Janouskova, and Jiri Matas.
\newblock Flaws of imagenet, computer vision's favourite dataset.
\newblock \emph{arXiv preprint arXiv:2412.00076}, 2024.

\bibitem[Krizhevsky et~al.(2012)Krizhevsky, Sutskever, and Hinton]{krizhevsky2012imagenet}
Alex Krizhevsky, Ilya Sutskever, and Geoffrey~E Hinton.
\newblock Imagenet classification with deep convolutional neural networks.
\newblock \emph{Advances in neural information processing systems}, 25, 2012.

\bibitem[Li et~al.(2018)Li, Pan, Wang, and Kot]{li2018domain}
Haoliang Li, Sinno~Jialin Pan, Shiqi Wang, and Alex~C Kot.
\newblock Domain generalization with adversarial feature learning.
\newblock In \emph{Proceedings of the IEEE conference on computer vision and pattern recognition}, pages 5400--5409, 2018.

\bibitem[Lin et~al.(2023)Lin, Tan, Hao, Wong, Dong, Zhang, Yang, and Zhang]{lin2023spurious}
Yong Lin, Lu Tan, Yifan Hao, Honam Wong, Hanze Dong, Weizhong Zhang, Yujiu Yang, and Tong Zhang.
\newblock Spurious feature diversification improves out-of-distribution generalization.
\newblock \emph{arXiv preprint arXiv:2309.17230}, 2023.

\bibitem[Liu et~al.(2021)Liu, Haghgoo, Chen, Raghunathan, Koh, Sagawa, Liang, and Finn]{liu2021just}
Evan~Z Liu, Behzad Haghgoo, Annie~S Chen, Aditi Raghunathan, Pang~Wei Koh, Shiori Sagawa, Percy Liang, and Chelsea Finn.
\newblock Just train twice: Improving group robustness without training group information.
\newblock In \emph{International Conference on Machine Learning}, pages 6781--6792. PMLR, 2021.

\bibitem[Liu et~al.(2023)Liu, Zeng, Ren, Li, Zhang, Yang, Li, Yang, Su, Zhu, et~al.]{liu2023grounding}
Shilong Liu, Zhaoyang Zeng, Tianhe Ren, Feng Li, Hao Zhang, Jie Yang, Chunyuan Li, Jianwei Yang, Hang Su, Jun Zhu, et~al.
\newblock Grounding dino: Marrying dino with grounded pre-training for open-set object detection.
\newblock \emph{arXiv preprint arXiv:2303.05499}, 2023.

\bibitem[Liu et~al.()]{convnextv2}
Zhuang Liu et~al.
\newblock Convnext-v2.
\newblock \url{https://github.com/facebookresearch/ConvNeXt-V2}.
\newblock Accessed: 2025-05-09.

\bibitem[Luo et~al.(2023)Luo, Pan, Sun, Zhang, Xiong, and Wu]{luo2023camouflaged}
Naisong Luo, Yuwen Pan, Rui Sun, Tianzhu Zhang, Zhiwei Xiong, and Feng Wu.
\newblock Camouflaged instance segmentation via explicit de-camouflaging.
\newblock In \emph{Proceedings of the IEEE/CVF conference on computer vision and pattern recognition}, pages 17918--17927, 2023.

\bibitem[Lynch et~al.(2023)Lynch, Dovonon, Kaddour, and Silva]{lynch2023spawrious}
Aengus Lynch, Gbètondji J-S Dovonon, Jean Kaddour, and Ricardo Silva.
\newblock Spawrious: A benchmark for fine control of spurious correlation biases, 2023.

\bibitem[maintainers and contributors(2016)]{torchvision2016}
TorchVision maintainers and contributors.
\newblock Torchvision: Pytorch's computer vision library.
\newblock \url{https://github.com/pytorch/vision}, 2016.

\bibitem[Minderer et~al.(2022)Minderer, Gritsenko, Stone, Neumann, Weissenborn, Dosovitskiy, Mahendran, Arnab, Dehghani, Shen, et~al.]{minderer2022simple}
Matthias Minderer, Alexey Gritsenko, Austin Stone, Maxim Neumann, Dirk Weissenborn, Alexey Dosovitskiy, Aravindh Mahendran, Anurag Arnab, Mostafa Dehghani, Zhuoran Shen, et~al.
\newblock Simple open-vocabulary object detection.
\newblock In \emph{European Conference on Computer Vision}, pages 728--755. Springer, 2022.

\bibitem[Minderer et~al.(2024)Minderer, Gritsenko, and Houlsby]{minderer2024scaling}
Matthias Minderer, Alexey Gritsenko, and Neil Houlsby.
\newblock Scaling open-vocabulary object detection.
\newblock \emph{Advances in Neural Information Processing Systems}, 36, 2024.

\bibitem[Moayeri et~al.(2022{\natexlab{a}})Moayeri, Pope, Balaji, and Feizi]{moayeri2022comprehensive}
Mazda Moayeri, Phillip Pope, Yogesh Balaji, and Soheil Feizi.
\newblock A comprehensive study of image classification model sensitivity to foregrounds, backgrounds, and visual attributes.
\newblock In \emph{Proceedings of the IEEE/CVF Conference on Computer Vision and Pattern Recognition}, pages 19087--19097, 2022{\natexlab{a}}.

\bibitem[Moayeri et~al.(2022{\natexlab{b}})Moayeri, Singla, and Feizi]{moayeri2022hard}
Mazda Moayeri, Sahil Singla, and Soheil Feizi.
\newblock Hard imagenet: Segmentations for objects with strong spurious cues.
\newblock \emph{Advances in Neural Information Processing Systems}, 35:\penalty0 10068--10077, 2022{\natexlab{b}}.

\bibitem[Noohdani et~al.(2024)Noohdani, Hosseini, Parast, Araghi, and Baghshah]{noohdani2024decompose}
Fahimeh~Hosseini Noohdani, Parsa Hosseini, Aryan~Yazdan Parast, Hamidreza~Yaghoubi Araghi, and Mahdieh~Soleymani Baghshah.
\newblock Decompose-and-compose: A compositional approach to mitigating spurious correlation.
\newblock In \emph{Proceedings of the IEEE/CVF Conference on Computer Vision and Pattern Recognition}, pages 27662--27671, 2024.

\bibitem[Oliva and Torralba(2007)]{oliva2007role}
Aude Oliva and Antonio Torralba.
\newblock The role of context in object recognition.
\newblock \emph{Trends in cognitive sciences}, 11\penalty0 (12):\penalty0 520--527, 2007.

\bibitem[Oquab et~al.(2023)Oquab, Darcet, Moutakanni, Vo, Szafraniec, Khalidov, Fernandez, Haziza, Massa, El-Nouby, et~al.]{oquab2023dinov2}
Maxime Oquab, Timoth{\'e}e Darcet, Th{\'e}o Moutakanni, Huy Vo, Marc Szafraniec, Vasil Khalidov, Pierre Fernandez, Daniel Haziza, Francisco Massa, Alaaeldin El-Nouby, et~al.
\newblock Dinov2: Learning robust visual features without supervision.
\newblock \emph{arXiv preprint arXiv:2304.07193}, 2023.

\bibitem[Picek et~al.(2022)Picek, \v{S}ulc, Matas, Jeppesen, Heilmann-Clausen, L{\ae}ss{\o}e, and Fr{\o}slev]{Picek_2022_WACV}
Luk\'a\v{s} Picek, Milan \v{S}ulc, Ji\v{r}{\'\i} Matas, Thomas~S. Jeppesen, Jacob Heilmann-Clausen, Thomas L{\ae}ss{\o}e, and Tobias Fr{\o}slev.
\newblock Danish fungi 2020 - not just another image recognition dataset.
\newblock In \emph{Proceedings of the IEEE/CVF Winter Conference on Applications of Computer Vision (WACV)}, pages 1525--1535, 2022.

\bibitem[Picek et~al.(2024{\natexlab{a}})Picek, Janouskova, Sulc, and Matas]{picek2024fungitastic}
Lukas Picek, Klara Janouskova, Milan Sulc, and Jiri Matas.
\newblock Fungitastic: A multi-modal dataset and benchmark for image categorization.
\newblock \emph{arXiv preprint arXiv:2408.13632}, 2024{\natexlab{a}}.

\bibitem[Picek et~al.(2024{\natexlab{b}})Picek, Neumann, and Matas]{picek2024animal}
Lukas Picek, Luk{\'a}{\v{s}} Neumann, and Ji{\v{r}}{\'\i} Matas.
\newblock Animal identification with independent foreground and background modeling.
\newblock In \emph{DAGM German Conference on Pattern Recognition}, pages 241--257. Springer, 2024{\natexlab{b}}.

\bibitem[Radford et~al.(2021)Radford, Kim, Hallacy, Ramesh, Goh, Agarwal, Sastry, Askell, Mishkin, Clark, et~al.]{radford2021learning}
Alec Radford, Jong~Wook Kim, Chris Hallacy, Aditya Ramesh, Gabriel Goh, Sandhini Agarwal, Girish Sastry, Amanda Askell, Pamela Mishkin, Jack Clark, et~al.
\newblock Learning transferable visual models from natural language supervision.
\newblock In \emph{International conference on machine learning}, pages 8748--8763. PMLR, 2021.

\bibitem[Ravi et~al.(2024)Ravi, Gabeur, Hu, Hu, Ryali, Ma, Khedr, R{\"a}dle, Rolland, Gustafson, et~al.]{ravi2024sam}
Nikhila Ravi, Valentin Gabeur, Yuan-Ting Hu, Ronghang Hu, Chaitanya Ryali, Tengyu Ma, Haitham Khedr, Roman R{\"a}dle, Chloe Rolland, Laura Gustafson, et~al.
\newblock Sam 2: Segment anything in images and videos.
\newblock \emph{arXiv preprint arXiv:2408.00714}, 2024.

\bibitem[Recht et~al.(2019)Recht, Roelofs, Schmidt, and Shankar]{recht2019imagenet}
Benjamin Recht, Rebecca Roelofs, Ludwig Schmidt, and Vaishaal Shankar.
\newblock Do imagenet classifiers generalize to imagenet?
\newblock In \emph{International conference on machine learning}, pages 5389--5400. PMLR, 2019.

\bibitem[Russakovsky et~al.(2015)Russakovsky, Deng, Su, Krause, Satheesh, Ma, Huang, Karpathy, Khosla, Bernstein, et~al.]{russakovsky2015imagenet}
Olga Russakovsky, Jia Deng, Hao Su, Jonathan Krause, Sanjeev Satheesh, Sean Ma, Zhiheng Huang, Andrej Karpathy, Aditya Khosla, Michael Bernstein, et~al.
\newblock Imagenet large scale visual recognition challenge.
\newblock \emph{International journal of computer vision}, 115:\penalty0 211--252, 2015.

\bibitem[Sagawa et~al.(2019)Sagawa, Koh, Hashimoto, and Liang]{sagawa2019distributionally}
Shiori Sagawa, Pang~Wei Koh, Tatsunori~B Hashimoto, and Percy Liang.
\newblock Distributionally robust neural networks for group shifts: On the importance of regularization for worst-case generalization.
\newblock \emph{arXiv preprint arXiv:1911.08731}, 2019.

\bibitem[Saranrittichai et~al.(2024)Saranrittichai, Munoz, Fischer, and Mummadi]{saranrittichai2024zeroshot}
Piyapat Saranrittichai, Mauricio Munoz, Volker Fischer, and Chaithanya~Kumar Mummadi.
\newblock Zero-shot recognition with guided cropping.
\newblock In \emph{ICLR 2024 Workshop on Mathematical and Empirical Understanding of Foundation Models}, 2024.

\bibitem[Shetty et~al.(2019)Shetty, Schiele, and Fritz]{shetty2019not}
Rakshith Shetty, Bernt Schiele, and Mario Fritz.
\newblock Not using the car to see the sidewalk--quantifying and controlling the effects of context in classification and segmentation.
\newblock In \emph{Proceedings of the IEEE/CVF Conference on Computer Vision and Pattern Recognition}, pages 8218--8226, 2019.

\bibitem[Shi et~al.(2021)Shi, Seely, Torr, Siddharth, Hannun, Usunier, and Synnaeve]{shi2021gradient}
Yuge Shi, Jeffrey Seely, Philip~HS Torr, N Siddharth, Awni Hannun, Nicolas Usunier, and Gabriel Synnaeve.
\newblock Gradient matching for domain generalization.
\newblock \emph{arXiv preprint arXiv:2104.09937}, 2021.

\bibitem[Singla and Feizi(2022)]{singla2022salient}
Sahil Singla and Soheil Feizi.
\newblock Salient imagenet: How to discover spurious features in deep learning?
\newblock In \emph{International Conference on Learning Representations}, 2022.

\bibitem[Stevens et~al.(2024)Stevens, Wu, Thompson, Campolongo, Song, Carlyn, Dong, Dahdul, Stewart, Berger-Wolf, et~al.]{stevens2024bioclip}
Samuel Stevens, Jiaman Wu, Matthew~J Thompson, Elizabeth~G Campolongo, Chan~Hee Song, David~Edward Carlyn, Li Dong, Wasila~M Dahdul, Charles Stewart, Tanya Berger-Wolf, et~al.
\newblock Bioclip: A vision foundation model for the tree of life.
\newblock In \emph{Proceedings of the IEEE/CVF conference on computer vision and pattern recognition}, pages 19412--19424, 2024.

\bibitem[Sun and Saenko(2016)]{sun2016deep}
Baochen Sun and Kate Saenko.
\newblock Deep coral: Correlation alignment for deep domain adaptation.
\newblock In \emph{Computer Vision--ECCV 2016 Workshops: Amsterdam, The Netherlands, October 8-10 and 15-16, 2016, Proceedings, Part III 14}, pages 443--450. Springer, 2016.

\bibitem[Taesiri et~al.(2024)Taesiri, Nguyen, Habchi, Bezemer, and Nguyen]{taesiri2024imagenet}
Mohammad~Reza Taesiri, Giang Nguyen, Sarra Habchi, Cor-Paul Bezemer, and Anh Nguyen.
\newblock Imagenet-hard: The hardest images remaining from a study of the power of zoom and spatial biases in image classification.
\newblock \emph{Advances in Neural Information Processing Systems}, 36, 2024.

\bibitem[Torralba(2003)]{torralba2003contextual}
Antonio Torralba.
\newblock Contextual priming for object detection.
\newblock \emph{International journal of computer vision}, 53:\penalty0 169--191, 2003.

\bibitem[Tschannen et~al.(2025)Tschannen, Gritsenko, Wang, Naeem, Alabdulmohsin, Parthasarathy, Evans, Beyer, Xia, Mustafa, et~al.]{tschannen2025siglip}
Michael Tschannen, Alexey Gritsenko, Xiao Wang, Muhammad~Ferjad Naeem, Ibrahim Alabdulmohsin, Nikhil Parthasarathy, Talfan Evans, Lucas Beyer, Ye Xia, Basil Mustafa, et~al.
\newblock Siglip 2: Multilingual vision-language encoders with improved semantic understanding, localization, and dense features.
\newblock \emph{arXiv preprint arXiv:2502.14786}, 2025.

\bibitem[Vapnik(1991)]{vapnik1991principles}
Vladimir Vapnik.
\newblock Principles of risk minimization for learning theory.
\newblock \emph{Advances in neural information processing systems}, 4, 1991.

\bibitem[Wang et~al.(2022)Wang, Machiraju, Choung, Herzog, and Frossard]{wang2022clad}
Ke Wang, Harshitha Machiraju, Oh-Hyeon Choung, Michael Herzog, and Pascal Frossard.
\newblock Clad: A contrastive learning based approach for background debiasing.
\newblock \emph{arXiv preprint arXiv:2210.02748}, 2022.

\bibitem[Wang et~al.(2025)Wang, Lin, Chen, Schmidt, Han, and Zhang]{wang2025sober}
Qizhou Wang, Yong Lin, Yongqiang Chen, Ludwig Schmidt, Bo Han, and Tong Zhang.
\newblock A sober look at the robustness of clips to spurious features.
\newblock \emph{Advances in Neural Information Processing Systems}, 37:\penalty0 122484--122523, 2025.

\bibitem[Wang et~al.(2023)Wang, Girdhar, Yu, and Misra]{wang2023cut}
Xudong Wang, Rohit Girdhar, Stella~X Yu, and Ishan Misra.
\newblock Cut and learn for unsupervised object detection and instance segmentation.
\newblock In \emph{Proceedings of the IEEE/CVF conference on computer vision and pattern recognition}, pages 3124--3134, 2023.

\bibitem[Wightman(2019)]{rw2019timm}
Ross Wightman.
\newblock Pytorch image models.
\newblock \url{https://github.com/rwightman/pytorch-image-models}, 2019.

\bibitem[Wolf et~al.(2020)Wolf, Debut, Sanh, Chaumond, Delangue, Moi, Cistac, Rault, Louf, Funtowicz, et~al.]{wolf2020transformers}
Thomas Wolf, Lysandre Debut, Victor Sanh, Julien Chaumond, Clement Delangue, Anthony Moi, Pierric Cistac, Tim Rault, R{\'e}mi Louf, Morgan Funtowicz, et~al.
\newblock Transformers: State-of-the-art natural language processing.
\newblock In \emph{Proceedings of the 2020 conference on empirical methods in natural language processing: system demonstrations}, pages 38--45, 2020.

\bibitem[Woo et~al.(2023)Woo, Debnath, Hu, Chen, Liu, Kweon, and Xie]{woo2023convnext}
Sanghyun Woo, Shoubhik Debnath, Ronghang Hu, Xinlei Chen, Zhuang Liu, In~So Kweon, and Saining Xie.
\newblock Convnext v2: Co-designing and scaling convnets with masked autoencoders.
\newblock In \emph{Proceedings of the IEEE/CVF Conference on Computer Vision and Pattern Recognition}, pages 16133--16142, 2023.

\bibitem[Xiao et~al.(2020)Xiao, Engstrom, Ilyas, and Madry]{xiao2020noise}
Kai Xiao, Logan Engstrom, Andrew Ilyas, and Aleksander Madry.
\newblock Noise or signal: The role of image backgrounds in object recognition.
\newblock \emph{arXiv preprint arXiv:2006.09994}, 2020.

\bibitem[Xu et~al.(2020)Xu, Zhang, Ni, Li, Wang, Tian, and Zhang]{xu2020adversarial}
Minghao Xu, Jian Zhang, Bingbing Ni, Teng Li, Chengjie Wang, Qi Tian, and Wenjun Zhang.
\newblock Adversarial domain adaptation with domain mixup.
\newblock In \emph{Proceedings of the AAAI conference on artificial intelligence}, pages 6502--6509, 2020.

\bibitem[Xu et~al.(2025)Xu, Xiang, and Liang]{xu2025overcoming}
Zhuo Xu, Xiang Xiang, and Yifan Liang.
\newblock Overcoming shortcut problem in vlm for robust out-of-distribution detection.
\newblock In \emph{Proceedings of the Computer Vision and Pattern Recognition Conference}, pages 15402--15412, 2025.

\bibitem[Yang et~al.(2024)Yang, Yang, Wu, and Feng]{yang2024significant}
Shengying Yang, Xinqi Yang, Jianfeng Wu, and Boyang Feng.
\newblock Significant feature suppression and cross-feature fusion networks for fine-grained visual classification.
\newblock \emph{Scientific Reports}, 14\penalty0 (1):\penalty0 24051, 2024.

\bibitem[Yao et~al.(2022)Yao, Wang, Li, Zhang, Liang, Zou, and Finn]{yao2022improving}
Huaxiu Yao, Yu Wang, Sai Li, Linjun Zhang, Weixin Liang, James Zou, and Chelsea Finn.
\newblock Improving out-of-distribution robustness via selective augmentation.
\newblock In \emph{International Conference on Machine Learning}, pages 25407--25437. PMLR, 2022.

\bibitem[Ye et~al.(2024)Ye, Zheng, Cao, Ma, and Zhang]{ye2024spurious}
Wenqian Ye, Guangtao Zheng, Xu Cao, Yunsheng Ma, and Aidong Zhang.
\newblock Spurious correlations in machine learning: A survey.
\newblock \emph{arXiv preprint arXiv:2402.12715}, 2024.

\bibitem[Zhao et~al.(2023)Zhao, Ding, An, Du, Yu, Li, Tang, and Wang]{zhao2023fast}
Xu Zhao, Wenchao Ding, Yongqi An, Yinglong Du, Tao Yu, Min Li, Ming Tang, and Jinqiao Wang.
\newblock Fast segment anything.
\newblock \emph{arXiv preprint arXiv:2306.12156}, 2023.

\bibitem[Zhu et~al.(2017)Zhu, Xie, and Yuille]{zhu2017object}
Zhuotun Zhu, Lingxi Xie, and Alan Yuille.
\newblock Object recognition with and without objects.
\newblock In \emph{Proceedings of the 26th International Joint Conference on Artificial Intelligence}, pages 3609--3615, 2017.

\bibitem[Zitnick and Doll{\'a}r(2014)]{zitnick2014edge}
C~Lawrence Zitnick and Piotr Doll{\'a}r.
\newblock Edge boxes: Locating object proposals from edges.
\newblock In \emph{Computer Vision--ECCV 2014: 13th European Conference, Zurich, Switzerland, September 6-12, 2014, Proceedings, Part V 13}, pages 391--405. Springer, 2014.

\end{thebibliography}
